\documentclass[sigconf]{acmart}
\usepackage{microtype}
\usepackage{graphicx}
\usepackage{subfigure}
\usepackage{booktabs} 
\usepackage{multirow} 
\usepackage{pifont}
\usepackage{bm}
\usepackage{enumitem}
\usepackage{algorithm}
\usepackage{algorithmic}

\usepackage{color} 

\usepackage{xcolor}
\usepackage{colortbl,booktabs}
\usepackage{wrapfig}
\usepackage{fontawesome5}

\newcommand{\cmark}{\ding{51}} 
\newcommand{\xmark}{\ding{55}}

\usepackage{hyperref}
\hypersetup{
	colorlinks=true,
	citecolor=cyan,
    linkcolor=red,
}

\usepackage{amsmath}
\usepackage{mathtools}
\usepackage{amsthm}

\AtBeginDocument{%
  }

\acmConference[MM '25]{Proceedings of the 33rd ACM International Conference on Multimedia (ACM MM 2025)}{October 27--31, 2025}{Dublin, Ireland}

\begin{document}

\title{DilateQuant: Accurate and Efficient Quantization-Aware Training for Diffusion Models via Weight Dilation}


\author{Xuewen Liu$^{1,2}$, \; Zhikai Li$^{1, }$\textsuperscript{\small\textcolor{black}{\faEnvelope}}, \; Minhao Jiang$^{1,2}$, \; Mengjuan Chen$^{1 \;}$, \; Jianquan Li$^{1 \;}$, \; Qingyi Gu$^{1, }$\textsuperscript{\small\textcolor{black}{\faEnvelope}}\\
\textnormal{$^1$Institute of Automation, Chinese Academy of Sciences\\
$^2$School of Artificial Intelligence, University of Chinese Academy of Sciences\\}
{\tt\small \{liuxuewen2023, lizhikai2020, qingyi.gu\}@ia.ac.cn}
}
\renewcommand{\shortauthors}{Xuewen Liu.}

\begin{abstract}
Model quantization is a promising method for accelerating and compressing diffusion models. 
Nevertheless, since post-training quantization (PTQ) fails catastrophically at low-bit cases, quantization-aware training (QAT) is essential.
Unfortunately, the wide range and time-varying activations in diffusion models sharply increase the complexity of quantization, making existing QAT methods inefficient.
Equivalent scaling can effectively reduce activation range, but previous methods remain the overall quantization error unchanged. More critically, these methods significantly disrupt the original weight distribution, resulting in poor weight initialization and challenging convergence during QAT training.
In this paper, we propose a novel QAT framework for diffusion models, called DilateQuant.
Specifically, we propose Weight Dilation (WD) that maximally dilates the unsaturated in-channel weights to a constrained range through equivalent scaling. 
WD decreases the activation range while preserving the original weight range, which steadily reduces the quantization error and ensures model convergence.
To further enhance accuracy and efficiency, we design a Temporal Parallel Quantizer (TPQ) to address the time-varying activations and introduce a Block-wise Knowledge Distillation (BKD) to reduce resource consumption in training.
Extensive experiments demonstrate that DilateQuant significantly outperforms existing methods in terms of accuracy and efficiency.
Code is available at {\textcolor{red}{\href{http://github.com/BienLuky/DilateQuant}{http://github.com/BienLuky/DilateQuant}}}
\end{abstract}
\vspace{-0.6cm}

\maketitle

\section{Introduction}

\begin{figure}[t]
    \centering
    \vspace{0.2cm}
    \includegraphics[width=0.92\linewidth]{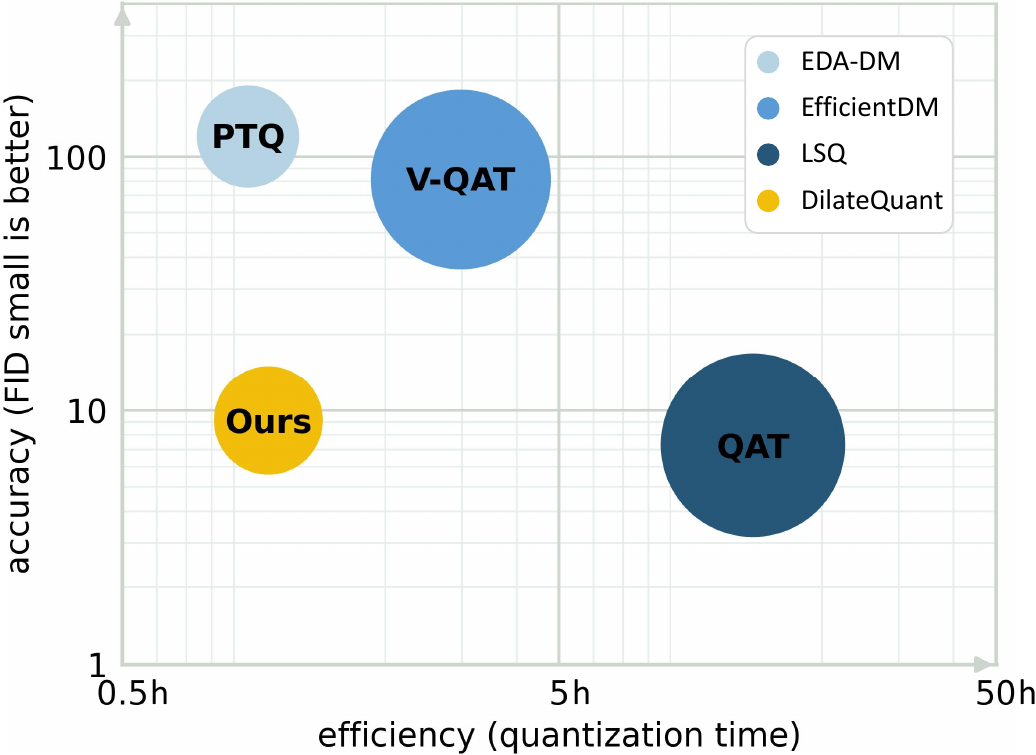}
    \vspace{-0.2cm}
    \caption{An overview of the accuracy-vs-efficiency trade-off across various approaches. The circle size represents GPU memory usage in quantization process. Data is collected from DDIM with 4-bit quantization on CIFAR-10.}
    \label{fig:efficient}
    \vspace{-0.5cm}
\end{figure}

Recently, diffusion models have shown excellent performance on visual generation~\citep{li2022srdiff,zhang2023adding,zhang2023inversion,ho2022imagen,wang2024lavie,xu2024imagereward,li2025k}, but the substantial computational costs and huge memory footprint hinder their low-latency applications in real-world scenarios.
Numerous methods~\citep{nichol2021improved,song2020denoising,lu2022dpm} have been proposed to find shorter sampling trajectories for the thousand iterations of the denoising process, effectively reducing latency.
However, complex networks with a large number of parameters used in each denoising step are computational and memory intensive, which slow down inference and consume high memory footprint.
For instance, the Stable-Diffusion~\citep{rombach2022high} with 16GB of running memory still takes over one second to perform one denoising step, even on the high-performance A6000.

Model quantization represents the weights and activations with low-bit integers, reducing memory requirements and accelerating computational operations. 
It is a highly promising way to facilitate the applications of diffusion models on source-constrained hardware.
For example, employing 8-bit models can achieve up to a 4$\times$ memory compression and 2.35$\times$ speedup compared to 32-bit full-precision models on a T4 GPU~\citep{kim2022integer}. 
Furthermore, adopting 4-bit models can deliver an additional 2$\times$ compression and 1.59$\times$ speedup compared to 8-bit models.

Typically, existing quantization techniques are implemented through two main approaches: Post-Training Quantization (PTQ) and Quantization-Aware Training (QAT).
As shown in Figure~\ref{fig:efficient}, PTQ~\citep{liu2024enhanced} calibrates the quantization parameter with a small calibration and does not rely on end-to-end retraining, making it data- and time-efficient.
However, it brings severe performance degradation at low bit-width.
In contrast, QAT~\citep{esser2019learned} can maintain performance by retraining the whole model, making it more desired for low-bit diffusion models.

\begin{figure*}[!t]
    \centering
    \includegraphics[width=0.96\textwidth]{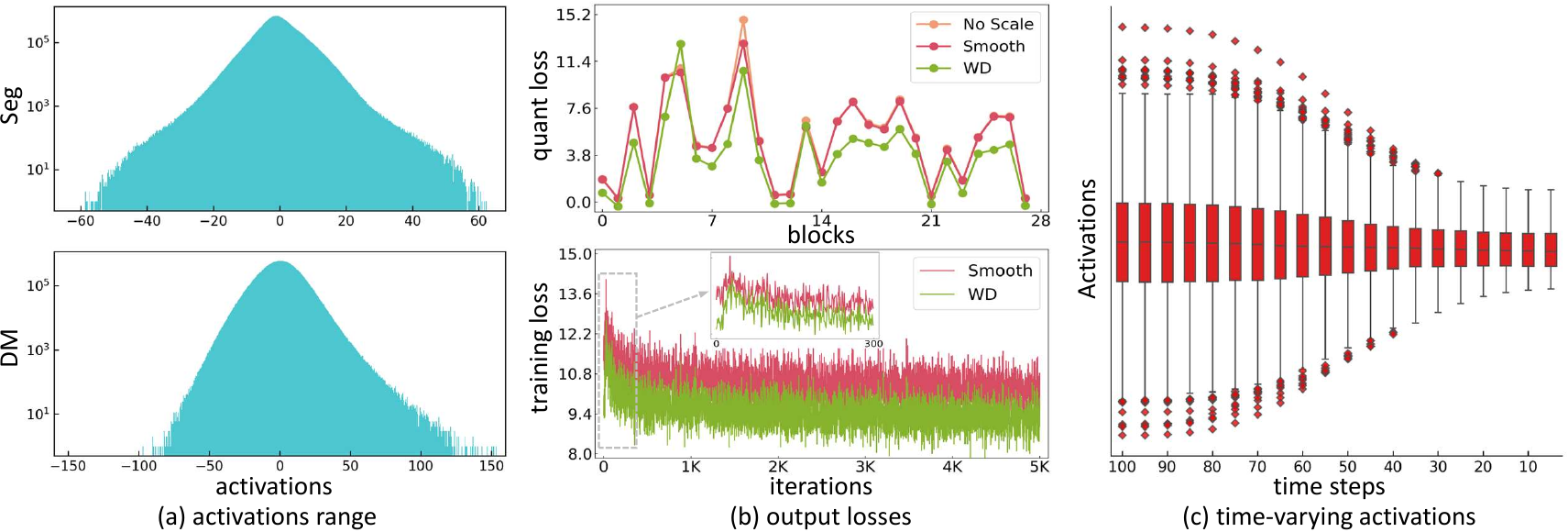}
    \vspace{-0.2cm}
    \caption{(a) showcases a wider range of activations in diffusion model (DM) compared to segmentation model (Seg). (b) demonstrates previous scaling methods are unsuitable for QAT of DM. (c) shows the dynamic distribution activations of DM. The activations of DM and Seg are from the first block output of the upsample stage of UNet network. The quant loss denotes the quantization errors of models without training and the training loss comes from the same block as (a).}
    \label{fig:distribution}
    \vspace{-0.2cm}
\end{figure*}

Unfortunately, standard QAT~\citep{esser2019learned} is impractical due to its time-consuming and resource-intensive nature. 
For instance, when applying both standard approaches to DDIM~\citep{song2020denoising} on CIFAR-10, QAT~\citep{esser2019learned} results in a 3.3$\times$ increase in GPU memory footprint (9.97 GB vs. 3.01 GB) and an 14.3$\times$ extension of quantization time (13.89 GPU-hours vs. 0.97 GPU-hours) compared to PTQ~\citep{liu2024enhanced}.
Although some variants of QAT~\cite{he2023efficientdm,wang2024quest} attempt to balance accuracy and efficiency, their performance remains unsatisfactory. The reason is primarily because the wide range and time-varying activations in diffusion models sharply increase the complexity of quantization.
Specifically, since the diffusion models infer in pixel space or latent space, the absence of layer normalization results in a wide range of activations.
For example, in the same UNet network, the range of activations is almost 2.5$\times$ larger than that of the segmentation models~\citep{ronneberger2015u}, as shown in Figure~\ref{fig:distribution}~(a).
Equivalent scaling can mitigate the wide range of activations. Previous methods~\citep{shao2023omniquant,lin2024awq,zhang2023dual,wei2023outlier}, particularly SmoothQuant~\citep{xiao2023smoothquant}, have proven effective in PTQ of large language models (LLMs). However, these methods are unsuitable for QAT of diffusion models. 
As illustrated in Figure~\ref{fig:distribution} (b), although these methods reduce the activation range, the overall quantization error remains unchanged due to the use of aggressive scaling factors. More importantly, the original weight distribution is significantly disrupted, resulting in poor weight initialization and challenging convergence during QAT training.
In addition, as shown in Figure~\ref{fig:distribution} (c), the temporal network induces a highly dynamic distribution of activations that varies across time steps, further diminishing the performance of quantization.

To address these issues,
we propose DilateQuant, a novel QAT framework that can achieve QAT-like accuracy with PTQ-like efficiency.
Specifically, we propose a weight-aware equivalent scaling algorithm, called Weight Dilation (WD), which searches for unsaturated in-channel weights and dilates them to the boundary of the quantized range.
By narrowing the activation range while keeping the weight range unchanged, WD steadily reduces the overall quantization errors and ensures model convergence during QAT training, mitigating the wide activation range in diffusion models.
To address time-varying activations, previous methods~\citep{he2023efficientdm,wang2024quest,huang2024tfmq} set multiple quantization parameters for one layer and trains them individually across time steps, which is data- and time-inefficient. 
In contrast, we design a unified Temporal Parallel Quantizer (TPQ), which supports parallel quantization using time-step quantization parameters through an indexing approach.
Additionally, inspired by PTQ reconstruction~\citep{li2021brecq}, we introduce Block-wise Knowledge Distillation (BKD), which distills the full-precision model into the quantized model using shorter backpropagation paths. 
TPQ and BKD further enhance accuracy, particularly by significantly improving efficiency, as evidenced by a 160$\times $ reduction in calibration overhead, a 3$\times $ reduction in GPU memory usage, and a 2$\times $ reduction in training time compared to the state-of-the-art method~\citep{he2023efficientdm} for DDIM on CIFAR-10.
The reproduction of DilateQuant is robust and easy as no hyper-parameters are introduced.
Overall, the contributions of this paper are as follows:
\begin{itemize}[leftmargin=*, itemsep=5pt, parsep=0pt]
    \vspace{-0.1cm}
    \item We formulate a novel QAT framework for diffusion models, DilateQuant, which offers comparable accuracy and high efficiency.
    \item The WD effectively alleviates the wide activation range in diffusion models. The TPQ and BKD further enhance performance while maintaining training efficiency.
    \item Through extensive experiments, we demonstrate that DilateQuant outperforms existing methods across lower quantization settings (6-bit, 4-bit), various models (DDPM, LDM-4, LDM-8, Stable-Diffusion, DiT-XL/2), and different datasets (CIFAR-10, LSUN-Bedroom, LSUN-Church, ImageNet, MS-COCO, DrawBench). 
\end{itemize}
\vspace{-0.2cm}

\section{Related Work}
\subsection{Diffusion Model Acceleration}
Diffusion models~\cite{peebles2023scalable,ho2020denoising,song2019generative,niu2020permutation} have generated high-quality images, but the substantial computational costs and huge memory footprint hinder their low-latency applications in real-world scenarios.
To reduce the inference computation, numerous methods have been proposed to find shorter sampling trajectories.
For example, \citep{nichol2021improved} shortens the denoising steps by adjusting variance schedule; \citep{song2020denoising,zhang2022gddim} generalize diffusion process to a non-Markovian process by modifying denoising equations; \citep{lu2022dpm,watson2022learning} use high-order solvers to approximate diffusion generation;
\citep{ma2024deepcache,chen2024delta,wimbauer2024cache} employ cache mechanism to reduce the inference path at each step.
These methods have achieved significant success.
Conversely, we focus on the complex networks of diffusion models, accelerating them at each denoising step with a quantization method, which not only reduces the computational cost but also compresses the model size. 

\begin{figure*}[!t]
    \centering
    \includegraphics[width=1.0\textwidth]{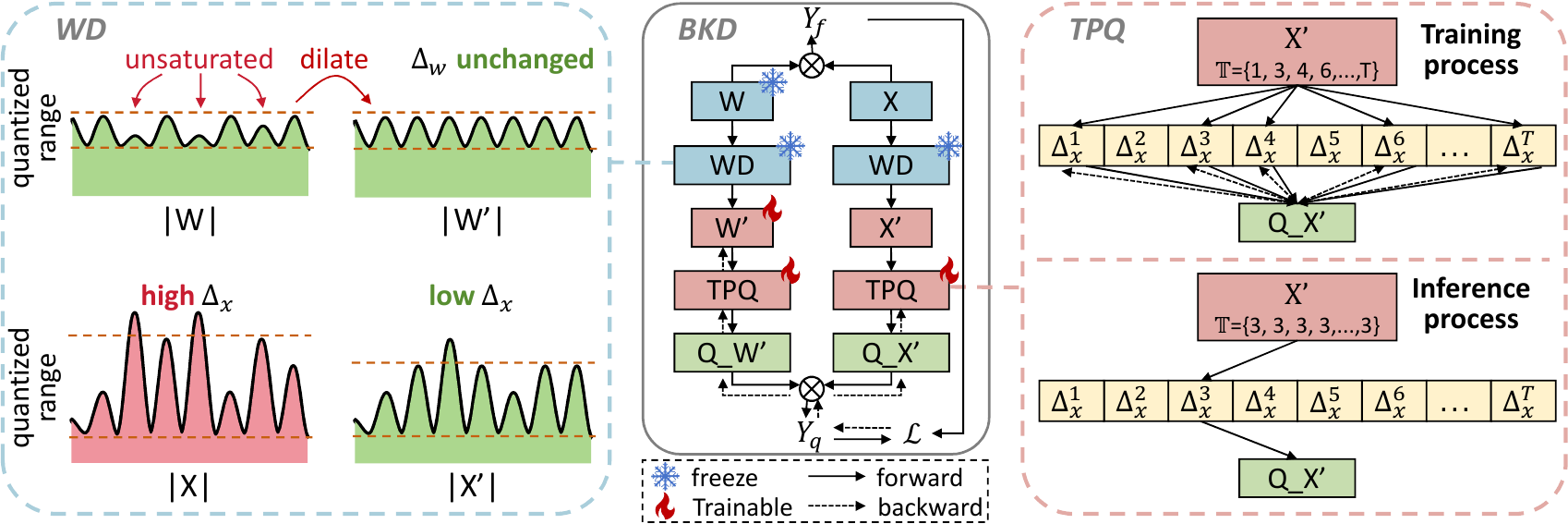}
    \caption{An overview of DilateQuant. WD narrows the activations range while maintaining the weights range unchanged. TPQ sets time-step quantization parameters and supports parallel training. BKD aligns the quantized network with the full-precision network at block level.}
    \label{fig:overview}
    \vspace{-0.1cm}
\end{figure*}

\subsection{Model Quantization}
Model quantization, which represents the original full-precision (FP) parameters with low-bit values, compresses model size and accelerates inference.
Depending on whether the model's weights are fine-tuned or not, it generally falls into two categories: Post-Training Quantization (PTQ)~\citep{nagel2020up,nagel2019data,shomron2021post,xiao2023patch,xiao2023dcifpn,li2023repq,li2024repquant} and Quantization-Aware Training (QAT)~\citep{louizos2018relaxed,jacob2018quantization,gong2019differentiable,li2023vit}.
PTQ calibrates quantization parameters with a small dataset and does not require fine-tuning the model's weights.
PTQ4DM~\citep{shang2023post} and Q-diffusion~\citep{li2023q} design specialized calibration and apply reconstruction~\citep{li2021brecq} to diffusion models. 
PTQD~\citep{he2023ptqd} uses statistical methods to estimate the quantization error.
TFMQ-DM~\citep{huang2024tfmq} changes the reconstruction module to align temporal information.
EDA-DM~\citep{liu2024enhanced} refines the reconstruction loss to avoid overfitting. 
TCAQ-DM~\citep{huang2024tcaq} employs a dynamically adaptive quantization module that mitigates the quantization error.
CacheQuant~\citep{liu2025cachequant} jointly optimizes quantization and caching techniques to achieve a higher acceleration ratio.
Although these PTQ methods enhance results, none of them break through the 6-bit quantization.
SVDQuant~\citep{li2024svdqunat} performs low-rank decomposition on outliers to achieve 4-bit quantization. However, it introduces additional computations and requires specialized operator support.
On the other hand, QAT retrains the whole model after the quantization operation, maintaining performance at lower bit-width. 
However, the significant training resources make it not practical for diffusion models.
For instance, TDQ~\citep{so2023temporal} requires 200K training iterations on a 50K original dataset.
To balance quantization accuracy and efficiency, some variants of QAT have been proposed.
EfficientDM~\citep{he2023efficientdm} fine-tunes all of the model's weights with an additional LoRA module, while QuEST~\citep{wang2024quest} selectively trains some sensitive layers.
Unfortunately, although they achieve 4-bit quantization of the diffusion models, both of them are non-standard (please refer to Appendix E for detail).
BitsFusion~\citep{sui2024bitsfusion} also incurs high computational costs by training the whole model to achieve 1.99-bit mixed-precision weight-only quantization.
Hence, the standard quantization of low-bit diffusion models with high accuracy and efficiency is still an open question.

\section{Preliminaries}

\subsection{Quantization}
Uniform quantizer is one of the most hardware-friendly choices, and we use it in our work.
The quantization-dequantization process of it can be defined as:
\begin{align}
    Quant&: x_{int}=clip \left( \left\lfloor\frac{x}{\Delta }\right\rceil+z, 0, 2^b-1 \right) \\
    DeQuant&: \hat{x}=\Delta \cdot  \left( x_{int}-z \right) \approx x
\end{align}
where $x$ and $x_{int}$ are the floating-point and quantized values, respectively, $\left\lfloor \cdot \right\rceil$ represents the rounding function, and the bit-width $b$ determines the range of clipping function $clip(\cdot)$.
In the dequantization process, the dequantized value $\hat{x}$ approximately recovers $x$. 
Notably, the theoretical derivations and experimental implementations in this paper are based on asymmetric quantization, in which the upper and lower bounds of $x$ determine the quantization parameters: scale factor $\Delta $ and zero-point $z$, as follows:
\begin{align}\label{eq:delta}
    \Delta  = \frac{max(x) - min(x)}{ 2^b-1 }, \quad z = \left\lfloor\frac{-min(x)}{\Delta }\right\rceil
\end{align}
Combining the two processes, we can provide a general definition for the quantization function, $Q(x)$, as:
\begin{align}\label{eq:quant}
    Q(x) = \Delta \cdot  \left( clip \left( \left\lfloor\frac{x}{\Delta }\right\rceil+z, 0, 2^b-1 \right)-z \right)
\end{align}
As can be seen, quantization is the process of introducing errors: $\left\lfloor \cdot \right\rceil$ and $clip(\cdot)$ result in rounding error ($\mathit{E_{round}}$) and clipping error ($\mathit{E_{clip}}$), respectively.
To set the quantization parameters, we commonly use two calibration methods: Max-Min and MSE.
For the former, quantization parameters are calibrated by the max-min values of $x$, eliminating the $\mathit{E_{clip}}$, but resulting in the largest $\Delta$; for the latter, quantization parameters are calibrated with appropriate values, but introduce the $\mathit{E_{clip}}$.

\subsection{Equivalent Scaling}
Equivalent scaling is a mathematically per-channel scaling transformation.
For a linear layer in diffusion models, the output $\bm{Y} = \bm{X} \bm{W}$, $\bm{Y} \in \mathbb{R}^{N\times C_o}$, $\bm{X} \in \mathbb{R}^{N\times C_i}$, $\bm{W} \in \mathbb{R}^{C_i\times C_o}$, where $N$ is the batch-size, $C_i$ is the in-channel, and $C_o$ is the out-channel. 
The activation $\bm{X}$ divides a per-in-channel scaling factor $\bm{s} \in \mathbb{R}^{C_i}$, and weight $\bm{W}$ scales accordingly in the reverse direction to maintain mathematical equivalence:
\begin{align}
    \bm{Y} = (\bm{X} / \bm{s}) (\bm{s} \cdot \bm{W})
\end{align}
The formula also suits the conv layer.
By ensuring that $\bm{s} > \bm{1}$, the range of activations can be made smaller and the range of weights larger, thus in transforming the difficulty of quantization from activations to weights.
In addition, given that the $\bm{X}$ is usually produced from previous linear operations, we can easily fuse the scaling factor into previous layers’ parameters offline so as not to introduce additional computational overhead in inference.
While some scaling methods~\citep{xiao2023smoothquant,lin2024awq,shao2023omniquant,zhang2023dual,wei2023outlier} have achieved success in PTQ framework of LLMs, they fail in QAT framework of diffusion models due to different quantization challenges, please see Appendix H for details.
\section{Methodology}

\subsection{Weight Dilation}

\begin{figure*}[!t]
    \centering
    \includegraphics[width=1.0\textwidth]{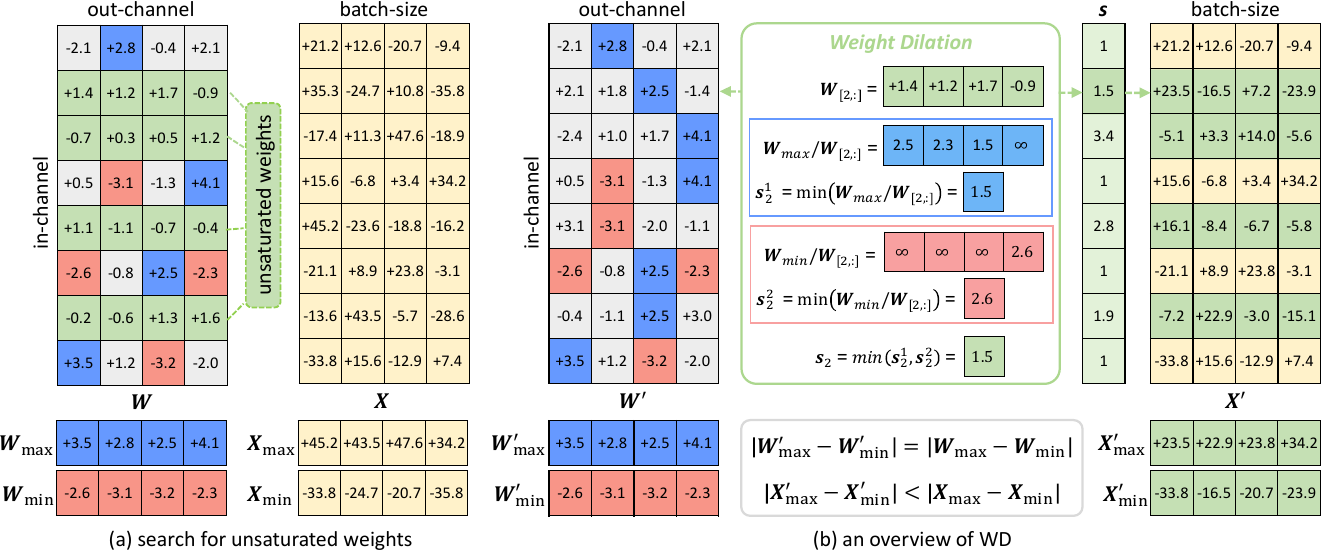}
    \vspace{-0.5cm}
    \caption{(a) WD searches for unsaturated in-channel weights based on the max-min values of each out-channel of the weights. (b) WD narrows the activation range by dilating unsaturated weights to a constrained range.}
    \label{fig:2}
    \vspace{-0.2cm}
\end{figure*}

\textbf{Analyzing Quantization Error.}
We start by analyzing the error from weight-activation quantization.
Considering that we calibrate the quantization parameters of activations and weights with a MSE manner and the zero-point \( z \) does not affect the quantization error before and after scaling, the quantization function (Eq.~\ref{eq:quant}) for $\bm{X}$ and $\bm{W}$ can be briefly written as:
\begin{align}
    Q(\bm{X}) &= \Delta_x \cdot  clip \left( \left\lfloor\frac{\bm{X}}{\Delta_x }\right\rceil \right) \label{eq:error1}\\
    Q(\bm{W}) &= \Delta_w \cdot  clip \left( \left\lfloor\frac{\bm{W}}{\Delta_w }\right\rceil \right) \label{eq:error2}
\end{align}
where $\Delta_x$ and $\Delta_w$ are scale factors. Thus, the quantization error can be defined as:
\begin{align}
    E(\bm{X}, \bm{W})=\|\bm{X W}-Q(\bm{X}) Q(\bm{W})\|_1
\end{align}
here, $\| \cdot \|_1$ denotes $L_1$ Norm. The formula can be further decomposed as:
\begin{equation}
    \begin{split}
    E& (\bm{X}, \bm{W}) \leq \|\bm{X}\|_1\|\bm{W}-Q(\bm{W})\|_1+ \\
    & \|\bm{X}-Q(\bm{X})\|_1\left(\|\bm{W}\|_1+\|\bm{W}-Q(\bm{W})\|_1\right)
    \end{split}
\end{equation}
Please see Appendix A for proof. Ultimately, the quantization error is influenced by four elements--the magnitude of the weight and activation, $\|\bm{W}\|_1$ and $\|\bm{X}\|_1$, and their respective quantization errors, $\|\bm{W}-Q(\bm{W})\|_1$ and $\|\bm{X}-Q(\bm{X})\|_1$.
Furthermore, the quantization errors result from rounding error ($\mathit{E_{round}}$) and cliping error ($\mathit{E_{clip}}$).
Given that $\mathit{E_{clip}}$ is negligibly small (as shown in Appendix F), the quantization errors can be expressed as:
\begin{align}
    \|\bm{X}-Q(\bm{X})\|_1 &= \Delta_x \cdot \mathit{E_{x_{round}}} \\
    \|\bm{W}-Q(\bm{W})\|_1 &= \Delta_w \cdot \mathit{E_{w_{round}}}
\end{align}
Since the rounding function maps a floating-point number to an integer, $\mathit{E_{round}}$ does not vary, as demonstrated in AWQ~\citep{lin2024awq}.
Previous methods (Smoothquant~\citep{xiao2023smoothquant}, OS+~\citep{wei2023outlier}, and Omniquant~\citep{shao2023omniquant}) scale the $\bm{X}$ and $\bm{W}$ using a \textbf{aggressive scaling factor} $\bm{s} \in \mathbb{R}^{C_i}$ to obtain the scaled $\bm{X}^{'}$ and $\bm{W}^{'}$, addressing outliers \textbf{in certain channels} of LLMs.
The quantization functions after scaling are:
\begin{align}
    Q(\bm{X}^{'}) & = Q(\bm{X}/\bm{s}) = \Delta_x^{'} \cdot  clip \left( \left\lfloor\frac{\bm{X}/\bm{s}}{\Delta_x^{'} }\right\rceil \right) \\
    Q(\bm{W}^{'}) & = Q(\bm{s} \cdot \bm{W}) = \Delta_w^{'} \cdot  clip \left( \left\lfloor\frac{\bm{s} \cdot \bm{W}}{\Delta_w^{'} }\right\rceil  \right)
\end{align}
where $\Delta_x^{'}$ and $\Delta_w^{'}$ are new scale factors. And the new magnitudes and quantization errors denote as:
\begin{align}
    \|\bm{X}^{'}\|_1 = \|\bm{X}\|_1 / \bm{s} &, \quad \|\bm{W}^{'}\|_1 = \bm{s} \cdot \|\bm{W}\|_1 \\
    \|\bm{X}-Q(\bm{X}^{'})\|_1& = \Delta_x^{'} \cdot \mathit{E_{x_{round}}} \\
    \|\bm{W}-Q(\bm{W}^{'})\|_1& = \Delta_w^{'} \cdot \mathit{E_{w_{round}}}
\end{align}
Given that outliers in diffusion models are present \textbf{across all channels} (as shown in Appendix Figure 8), applying aggressive scaling parameters results in $\Delta_x/\Delta_x^{'}=\Delta_w^{'}/\Delta_w=\bm{s}$. 
This implies that, although the $\|\bm{X}^{'}\|_1$ and $\|\bm{X}-Q(\bm{X}^{'})\|_1$ decrease while the $\|\bm{W}^{'}\|_1$ and $\|\bm{W}-Q(\bm{W}^{'})\|_1$ equivalently increasing.
Consequently, there are no overall change in $E(\bm{X}, \bm{W})$, as evidenced by the near-identical overlap between the quantization loss of SmoothQuant-scaled models and that of No-scaling, as shown in Figure~\ref{fig:distribution} (b).
There are also some methods (AWQ and DGQ~\citep{zhang2023dual}) that use \textbf{minimal scaling factors} to address outliers. However, these approaches are inadequate for handling the prevalent outliers.
We further validate these conclusions in Appendix H. 

\textbf{Analyzing Training Convergence.}
As a well-established fact, the original weight distribution significantly affects the training of deep neural networks, influencing both the training difficulty~\cite{hoedt2024principled} and performance ceiling~\cite{wong2024analysis}.
Given that QAT involves retraining the quantized weights and equivalent scaling alters the original weight distribution, it is crucial to carefully consider the impact of scaling on QAT training.
Previous scaling methods, such as SmoothQuant, employ aggressive scaling factors that significantly disrupt the original distribution of the pre-training weights.
We find that this results in poor weight initialization and challenging convergence during QAT training, severely impairing the performance of the quantized models.
Specifically, as shown in Figure~\ref{fig:distribution} (b), the poor weight initialization leads to larger quantization errors at the early stages of training, which slows down the convergence of models.
Moreover, the expanded weight range increases the likelihood of converging to a local optimum at the end stages of training, thereby reducing the performance ceiling of models.
PTQ4DiT~\citep{wu2024ptq4dit} firstly applies equivalent scaling into the PTQ framework of DiT models~\citep{peebles2023scalable,wang2024lavin}. However, it also employs aggressive scaling factors.

\textbf{Weight Dilation.}
Based on the above analysis, 
we propose Weight Dilation (\textbf{WD}), which is keenly aware of the unsaturated in-channel weights that can be cleverly exploited to reduce the range of activations.
Firstly, we model the problem of determining the scaling factors in a finer-grained manner:  
Considering that the quantization dimension of weights \( \bm{W} \) is per-out-channel, the quantization parameter for the \( j_{th} \) out-channel weights is given by:
\begin{align}
    \Delta_{w_j} = \frac{{max}(\bm{W}_{[:,j]}) - {min}(\bm{W}_{[:,j]})}{2^b-1 }
\end{align}
where $\Delta_{w_j}$ is the \( j_{th} \) elements of \( \Delta_w \in \mathbb{R}^{C_o} \). In contrast, the equivalent scaling is applied along the pre-in-channel dimension, with the scaling factor \( \bm{s} \in \mathbb{R}^{C_i} \). After scaling, the scaled activations and weights are \( \bm{X}' = \bm{X}/\bm{s} \) and \( \bm{W}' = \bm{s} \cdot \bm{W} \), respectively. The corresponding parameter becomes:
\begin{align}
    \Delta'_{w_j} = \frac{{max}(\bm{W}'_{[:,j]}) - {min}(\bm{W}'_{[:,j]})}{2^b-1 }
\end{align}
To maintain the original weight range while reducing the activation range, the constraints are as follows:  
\begin{equation}
    \begin{split}
    {max}(\bm{W}'_{[:,j]}) &= {max}(\bm{W}_{[:,j]}) \\ 
    {min}(\bm{W}'_{[:,j]}) &= {min}(\bm{W}_{[:,j]}) \\
    \|\bm{X}^{'}\|_1 &< \|\bm{X}\|_1
    \end{split}  
\end{equation}
This can be further expressed as \( \Delta'_w = \Delta_w \) and \( \bm{s} > 1 \).  
To minimize the activation range as much as possible, the final optimization objective is \( max( \bm{s} )\).  

Secondly, we solve the above problem as follows:
For the \( i_{th} \) element of the scaling factor $\bm{s}_i$, to ensure \( \Delta'_w = \Delta_w \), the scaled in-channel weights \( \bm{W}'_{[i,:]} = \bm{s}_i \cdot \bm{W}_{[i,:]} \) must satisfy:
\begin{equation}
    \begin{split}
    &\forall j\in \left \{  { 1,...,C_o}\right \}, \\
    &{min}(\bm{W}_{[:,j]}) \leq \bm{W}'_{[i,j]} \leq {max}(\bm{W}_{[:,j]})
    \end{split}
\end{equation}
Additionally, to satisfy \( \bm{s}_i > 1 \) and maximize \( \bm{s}_i \), WD dilates \( \bm{W}'_{[i,:]} \) to the boundary of the original weight range, i.e., $\bm{W}'_{[i,j]} = {max}(\bm{W}_{[:,j]})$ or $\bm{W}'_{[i,j]} = {min}(\bm{W}_{[:,j]})$.
Specifically, the solution is formulated as follows:
\begin{equation}\label{eq:solve}
    \begin{split}
    &\bm{s}^1_i = {\min_{\substack{j \in \{1, \dots, C_o\} \\ \bm{W}_{[i, j]} > 0}}} \left( {max}(\bm{W}_{[:,j]}) / \bm{W}_{[i,j]} \right) \\
    &\bm{s}^2_i = {\min_{\substack{j \in \{1, \dots, C_o\} \\ \bm{W}_{[i, j]} < 0}}} \left( {min}(\bm{W}_{[:,j]}) / \bm{W}_{[i,j]} \right) \\
    &max(\bm{s}_i) = \min(\bm{s}^1_i, \bm{s}^2_i)
    \end{split}
\end{equation}
here, $\bm{s}^1_i$ and $\bm{s}^2_i$ represent the maximized scaling factors constrained by \( {max}(\bm{W}_{[:,j]}) \) and \( {min}(\bm{W}_{[:,j]}) \), respectively, ${\min}$ ensure that $\bm{W}'_{[i,:]}$ do not exceed the range of all out-channels.
Given the symmetric distribution of weights where \( {max}(\bm{W}_{[:,j]}) > 0 \) and \( {min}(\bm{W}_{[:,j]}) < 0 \), we clamp \( W_{[i,j]} \) at \( \pm 1e^{-5} \) to avoid division by zero and sign changes.

Finally, WD searches for the unsaturated in-channel weights and dilates them to the boundary of the quantized range, narrowing the range of activations while keeping the weights range unchanged, as shown in Figure~\ref{fig:overview}.
More specifically, WD ensures the max-min values ($\bm{W}_{max} \in \mathbb{R}^{C_o}$, $\bm{W}_{min} \in \mathbb{R}^{C_o}$) of each out-channel unchanged and record their indexes of in-channel to form a set $\mathbb{A}$.
For example, the $\mathbb{A}$ in Figure~\ref{fig:2} (a) is \{1,4,6,8\}.
Iterating through the index of in-channel $k \in \left \{1,\dots ,C_i \right \}$, if $k \in \mathbb{A}$, we set $\bm{s}_k=1$, representing no scaling; if $k \notin \mathbb{A}$, the $\bm{W}_{[k,:]}$ denotes as unsaturated in-channel weights, and we set $\bm{s}_k$ by Eq.~\ref{eq:solve}.
As shown in Figure~\ref{fig:2} (b), WD reduces the range of activations while keeping the weight range unchanged.
Consequently, with $\|\bm{X}^{'}\|_1$, $\|\bm{X}-Q(\bm{X}^{'})\|_1$ reduced and $\|\bm{W}-Q(\bm{W}^{'})\|_1$ unchanged, WD steadily minimizes quantization errors. And the preserved original weight range ensures the convergence of QAT training. 
Furthermore, since unsaturated weights often correspond to extreme outliers (as demonstrated by PTQ4DiT), WD significantly reduces the activation range.
The workflow and effects of WD are detailed in Appendix G. 
\subsection{Temporal Parallel Quantizer}
Previous methods~\citep{he2023efficientdm,so2023temporal,wang2024quest} utilize multiple activation quantization parameters for one layer.
However, since each parameter is independent, these methods optimize each parameter individually, which is data- and time-inefficient.
For example, EfficientDM uses 819.2K samples for a total of 12.8K iterations for DDIM on CIFAR-10.

Different from these methods, as shown in Figure~\ref{fig:overview}, we design a novel quantizer, denotes as Temporal Parallel Quantizer (TPQ), which sets time-step quantization parameters for activations, instead of simply stacking parameters.
Specifically, in the QAT-training process, it utilizes \textit{an indexing approach} to call the corresponding parameters for the network inputs at different time steps. 
For a model with $T$ time steps, the quantization parameters of TPQ are as follows:
\begin{align}
    \Delta_x &= \left \{ \Delta_x^1, \Delta_x^2, \Delta_x^3, \dots, \Delta_x^{T} \right \} \\
    z_x &= \left \{ z_x^1, z_x^2, z_x^3, \dots, z_x^{T} \right \}
\end{align}
For linear and conv layers, they take inputs $\bm{x} \in \mathbb{R}^{|\mathbb{T}| \times C_i}$ and $\bm{x} \in \mathbb{R}^{|\mathbb{T}| \times  C_i \times H \times W}$, respectively, where $\mathbb{T}$ is a set containing different time-step indexes, $\mathbb{T} \subset \left \{ 1, \dots , T \right \}$, $\left | \cdot \right | $ represents the number of set elements.
The quantization operation of TPQ can be represented as:
\begin{align}\label{eq:quantizer}
    Q(\bm{x}) = \Delta_x^{\mathbb{T}} \cdot  \left( clip \left( \left\lfloor\frac{\bm{x}}{\Delta_x^{\mathbb{T}}}\right\rceil+z_x^{\mathbb{T}}\right)-z_x^{\mathbb{T}} \right)
\end{align}
where $\Delta_x^{\mathbb{T}}$ and $z_x^{\mathbb{T}}$ denote the quantization parameters corresponding to $\mathbb{T}$, respectively.
As shown in Figure~\ref{fig:overview}, taking the different time-step inputs as $X^{\mathbb{T}}$, where $\mathbb{T}=\left \{1,3,4,6 \right\}$, a single backward propagation can simultaneously update multiple quantization parameters ($\Delta_x^{\left \{1,3,4,6 \right\}}, z_x^{\left \{1,3,4,6 \right\}}$). In contrast to previous methods (TDQ~\cite{so2023temporal}, EfficientDM~\cite{he2023efficientdm}, QuEST~\cite{wang2024quest}), which train one parameter at a time, this parallel training of time-step parameters significantly reduces the data and time costs.
On the other hand, in the inference process, TPQ calls designated parameters for the network inputs at specific time steps, ensuring compatibility with standard CUDA operators. This eliminates the need for specialized operator designs and maintains deployment efficiency.

\subsection{Block-wise Knowledge Distillation}
Traditional QAT methods~\cite{esser2019learned,sui2024bitsfusion} alleviates accuracy degradation by end-to-end retraining of the whole complex networks, which is time- and memory-intensive. 
Assume the target model for quantization has $K$ blocks $\{B_1, \dots, B_K$\} with corresponding weights $w=\left \{ w_1, \dots, w_K\right\}$ and the input samples are $\bm{x}$, the training loss can be expressed as:
\begin{align}
    \mathcal{L}_{\bm{w}} = \mathcal{L} \left ( B_K(\bm{x})
    - \hat{B}_K(\bm{x}) \right ) 
\end{align}
where $\hat{B}_K(\bm{x})$ is the output of quantized model.
To improve training efficiency, reconstruction-based PTQ methods~\cite{nagel2020up,shao2023omniquant,ding2023cbq} optimize several quantization parameters block by block. For the target block $B_k$, the training loss is formulated as follows:
\begin{align}
    \mathcal{L}_{\bm{\theta }_k} = \mathcal{L} \left ( B_k(\bm{x})
    - \hat{B}_k(\bm{x}) \right ) 
\end{align}
where the trained parameters $\bm{\theta }_k$ for $k_{th}$ block can be step sizes~\cite{ding2023cbq}, clipping parameters~\cite{shao2023omniquant}, and rounding parameters~\cite{nagel2020up}.
Although these methods (PTQD~\cite{he2023ptqd}, EDA-DM~\citep{liu2024enhanced}, TFMQ-DM~\cite{huang2024tfmq}, TCAQ-DM~\citep{huang2024tcaq} and PTQ4DiT~\cite{wu2024ptq4dit}) ensure training efficiency, they fail to recover the accuracy loss of low-bit diffusion models.

To enhance performance while maintaining efficiency, inspired by the reconstruction method in PTQ~\citep{li2021brecq}, we propose a novel distillation strategy called Block-wise Knowledge Distillation (BKD).
BKD trains the quantized network block-by-block and simultaneously update the quantization parameters $(\Delta_x^{\mathbb{T}}, z_x^{\mathbb{T}}, \Delta_{w_k})$ and weights ($\bm{w}_k$) of $\hat{B}_k$ using the mean square loss $\mathcal{L}$:
\begin{align}
    \mathcal{L}_{\Delta_x^{\mathbb{T}}, z_x^{\mathbb{T}}, \Delta_{w_k}, \bm{w}_k} = MSE \left ( B_k(\bm{x})
    - \hat{B}_k(\bm{x}) \right ) 
\end{align}
As can be seen,
BKD retrains weights to recover accuracy and shortens the gradient backpropagation path to maintain efficiency.
In addition, BKD trains time-step quantization parameters and weights in parallel, which adapts the weights to each time step.

\begin{table}[t]\scriptsize 
    \centering
    \caption{Results of unconditional image generation. The “Calib.” presents the number of calibration and “Prec.” indicates the precision of weight and activation. \textsuperscript{$\star$} denotes our best implementation and \textsuperscript{$\dagger$} represents results directly obtained by rerunning open-source codes.}
    \vspace{-0.2cm}
    \label{tab:uncon}
    \setlength{\tabcolsep}{1.0mm}
    \resizebox{\linewidth}{!}{
    \begin{tabular}{ccccccc}
    \toprule 
    Task & Method & Calib. & Prec. & FID~$\downarrow$ & sFID~$\downarrow$ & IS~$\uparrow$ \\
    \cmidrule(r){1-7}
    \multirow{9}{1.35cm}{\centering \\CIFAR-10 \\32 $\times $ 32\\$ $\\DDPM\\steps = 100} & FP & - & 32/32 & 4.26 & 4.16 & 9.03 \\ 
    \cmidrule(r){2-7}
     & EDA-DM~\textsuperscript{$\dagger$} & 5120 & 6/6 & 26.68 & 14.10 & \textbf{9.35} \\
     & TFMQ-DM~\textsuperscript{$\dagger$} & 10240 & 6/6 & 9.59 & 7.84 & 8.84 \\
     & EfficientDM~\textsuperscript{$\star$} & 819.2K & 6/6 & 17.29 & 9.38 & 8.85 \\
     & \cellcolor[rgb]{0.843, 0.980, 0.816}DilateQuant 
     & \cellcolor[rgb]{0.843, 0.980, 0.816}5120 
     & \cellcolor[rgb]{0.843, 0.980, 0.816}6/6 
     & \cellcolor[rgb]{0.843, 0.980, 0.816}\textbf{4.46} 
     & \cellcolor[rgb]{0.843, 0.980, 0.816}\textbf{4.64} 
     & \cellcolor[rgb]{0.843, 0.980, 0.816}8.92 \\
    \cmidrule(r){2-7}
     & EDA-DM~\textsuperscript{$\dagger$} & 5120 & 4/4 & 120.24 & 36.72 & 4.42 \\
     & TFMQ-DM~\textsuperscript{$\dagger$} & 10240 & 4/4 & 236.63 & 59.66 & 3.19 \\
     & EfficientDM~\textsuperscript{$\star$} & 819.2K & 4/4 & 81.27 & 30.95 & 6.68 \\
     & \cellcolor[rgb]{0.843, 0.980, 0.816}DilateQuant 
     & \cellcolor[rgb]{0.843, 0.980, 0.816}5120 
     & \cellcolor[rgb]{0.843, 0.980, 0.816}4/4 
     & \cellcolor[rgb]{0.843, 0.980, 0.816}\textbf{9.13} 
     & \cellcolor[rgb]{0.843, 0.980, 0.816}\textbf{6.92} 
     & \cellcolor[rgb]{0.843, 0.980, 0.816}\textbf{8.56} \\
    \cmidrule(r){1-7}
    \multirow{11}{1.35cm}{\centering \\LSUN-Bedroom \\256 $\times $ 256\\$ $\\LDM-4\\steps = 100\\eta = 1.0} & FP & - & 32/32 & 3.02 & 7.21 & 2.29 \\ 
    \cmidrule(r){2-7}
     & EDA-DM~\textsuperscript{$\dagger$} & 5120 & 6/6 & 10.56 & 16.22 & 2.12 \\
     & TFMQ-DM~\textsuperscript{$\dagger$} & 10240 & 6/6 & 4.82 & 9.45 & 2.15 \\
     & EfficientDM~\textsuperscript{$\star$} & 102.4K & 6/6 & 5.43 & 15.11 & 2.15 \\
     & QuEST~\textsuperscript{$\star$} & 5120 & 6/6 & 10.1 & 19.57 & \textbf{2.20} \\
     & \cellcolor[rgb]{0.843, 0.980, 0.816}DilateQuant 
     & \cellcolor[rgb]{0.843, 0.980, 0.816}5120 
     & \cellcolor[rgb]{0.843, 0.980, 0.816}6/6 
     & \cellcolor[rgb]{0.843, 0.980, 0.816}\textbf{3.92} 
     & \cellcolor[rgb]{0.843, 0.980, 0.816}\textbf{8.90} 
     & \cellcolor[rgb]{0.843, 0.980, 0.816}2.17 \\
    \cmidrule(r){2-7}
     & EDA-DM~\textsuperscript{$\dagger$} & 5120 & 4/4 & N/A & N/A & N/A \\
     & TFMQ-DM~\textsuperscript{$\dagger$} & 10240 & 4/4 & 220.67 & 104.09 & N/A \\
     & EfficientDM~\textsuperscript{$\star$} & 102.4K & 4/4 & 15.27 & 19.87 & 2.11 \\
     & QuEST~\textsuperscript{$\star$} & 5120 & 4/4 & N/A & N/A & N/A \\
     & \cellcolor[rgb]{0.843, 0.980, 0.816}DilateQuant 
     & \cellcolor[rgb]{0.843, 0.980, 0.816}5120 
     & \cellcolor[rgb]{0.843, 0.980, 0.816}4/4 
     & \cellcolor[rgb]{0.843, 0.980, 0.816}\textbf{8.99} 
     & \cellcolor[rgb]{0.843, 0.980, 0.816}\textbf{14.88} 
     & \cellcolor[rgb]{0.843, 0.980, 0.816}\textbf{2.13} \\
    \cmidrule(r){1-7}
    \multirow{11}{1.35cm}{\centering \\LSUN-Church \\256 $\times $ 256\\$ $\\LDM-8\\steps = 100\\eta = 0.0} & FP & - & 32/32 & 4.06 & 10.89 & 2.70 \\ 
    \cmidrule(r){2-7}
     & EDA-DM~\textsuperscript{$\dagger$} & 5120 & 6/6 & 10.76 & 18.23 & 2.43 \\
     & TFMQ-DM~\textsuperscript{$\dagger$} & 10240 & 6/6 & 7.65 & 15.30 & \textbf{2.73} \\
     & EfficientDM~\textsuperscript{$\star$} & 102.4K & 6/6 & 6.92 & 12.84 & 2.65 \\
     & QuEST~\textsuperscript{$\star$} & 5120 & 6/6 & 6.83 & 11.93 & 2.65 \\
     & \cellcolor[rgb]{0.843, 0.980, 0.816}DilateQuant 
     & \cellcolor[rgb]{0.843, 0.980, 0.816}5120 
     & \cellcolor[rgb]{0.843, 0.980, 0.816}6/6 
     & \cellcolor[rgb]{0.843, 0.980, 0.816}\textbf{5.33} 
     & \cellcolor[rgb]{0.843, 0.980, 0.816}\textbf{11.61} 
     & \cellcolor[rgb]{0.843, 0.980, 0.816}2.66 \\
    \cmidrule(r){2-7}
     & EDA-DM~\textsuperscript{$\dagger$} & 5120 & 4/4 & N/A & N/A & N/A \\
     & TFMQ-DM~\textsuperscript{$\dagger$} & 10240 & 4/4 & 289.06 & 288.20 & 1.54 \\
     & EfficientDM~\textsuperscript{$\star$} & 102.4K & 4/4 & 15.08 & 16.53 & \textbf{2.67} \\
     & QuEST~\textsuperscript{$\star$} & 5120 & 4/4 & 13.03 & 19.50 & 2.63 \\
     & \cellcolor[rgb]{0.843, 0.980, 0.816}DilateQuant 
     & \cellcolor[rgb]{0.843, 0.980, 0.816}5120 
     & \cellcolor[rgb]{0.843, 0.980, 0.816}4/4 
     & \cellcolor[rgb]{0.843, 0.980, 0.816}\textbf{10.10} 
     & \cellcolor[rgb]{0.843, 0.980, 0.816}\textbf{16.22} 
     & \cellcolor[rgb]{0.843, 0.980, 0.816}2.62 \\
    \bottomrule
\end{tabular}
    }
    \vspace{-0.3cm}
\end{table}
\section{Experiment}
\subsection{Experimental Setup}
\paragraph{Models and Metrics.}
The comprehensive experiments include DDPM, LDM~\citep{song2020denoising,rombach2022high} and Stable-Diffusion on 5 datasets~\cite{lin2014microsoft,krizhevsky2009learning,yu2015lsun,deng2009imagenet}.
The performance of the quantized models is evaluated with FID~\citep{heusel2017gans}, sFID~\citep{salimans2016improved}, IS~\citep{salimans2016improved}, and CLIP score~\citep{hessel2021clipscore}.
Following the common practice, the Stable-Diffusion generates 10,000 images, while all other models generate 50,000 images. 
Besides, we extend DilateQuant to the DiT models~\citep{peebles2023scalable}, following~\cite{wu2024ptq4dit}, where the model generates 10,000 images for evaluation.

\begin{table}[!t]\footnotesize 
    \centering
    \caption{Results of class-conditional image generation.}
    \vspace{-0.2cm}
    \label{tab:con}
    \setlength{\tabcolsep}{1.0mm}
    \resizebox{\linewidth}{!}{
    \begin{tabular}{ccccccc}
    \toprule
    Task & Method & Calib. & Prec. & FID~$\downarrow$ & sFID~$\downarrow$ & IS~$\uparrow$ \\
    \cmidrule(r){1-7}
    \multirow{14}{1.35cm}{\centering \\ImageNet \\256 $\times $ 256 \\$ $ \\ LDM-4\\steps = 20\\eta=0.0\\scale = 3.0} & FP & - & 32/32 & 11.69 & 7.67 & 364.72 \\ 
    \cmidrule(r){2-7}
     & PTQD~\textsuperscript{$\dagger$} & 1024 & 6/6 & 16.38 & 17.79 & 146.78 \\
     & EDA-DM~\textsuperscript{$\dagger$} & 1024 & 6/6 & 11.52 & 8.02 & \textbf{360.77} \\
     & TFMQ-DM~\textsuperscript{$\dagger$} & 10240 & 6/6 & \textbf{7.83} & 8.23 & 311.32 \\
     & EfficientDM~\textsuperscript{$\star$} & 102.4K & 6/6 & 8.69 & 8.10 & 309.52 \\
     & QuEST~\textsuperscript{$\star$} & 5120 & 6/6 & 8.45 & 9.36 & 310.12 \\
     & \cellcolor[rgb]{0.843, 0.980, 0.816}DilateQuant 
     & \cellcolor[rgb]{0.843, 0.980, 0.816}1024 
     & \cellcolor[rgb]{0.843, 0.980, 0.816}6/6 
     & \cellcolor[rgb]{0.843, 0.980, 0.816}8.25 
     & \cellcolor[rgb]{0.843, 0.980, 0.816}\textbf{7.66} 
     & \cellcolor[rgb]{0.843, 0.980, 0.816}312.30 \\
    \cmidrule(r){2-7}
     & PTQD~\textsuperscript{$\dagger$} & 1024 & 4/4 & 245.84 & 107.63 & 2.88 \\
     & EDA-DM~\textsuperscript{$\dagger$} & 1024 & 4/4 & 20.02 & 36.66 & 204.93 \\
     & TFMQ-DM~\textsuperscript{$\dagger$} & 10240 & 4/4 & 258.81	& 152.42 & 2.40 \\
     & TCAQ-DM~\textsuperscript{$\dagger$} & - & 4/4 & 30.69 & 18.92 & 86.11 \\
     & EfficientDM~\textsuperscript{$\star$} & 102.4K & 4/4 & 12.08 & 14.75 & 122.12 \\
     & QuEST~\textsuperscript{$\star$} & 5120 & 4/4 & 38.43 & 29.27 & 69.58 \\
     & \cellcolor[rgb]{0.843, 0.980, 0.816}DilateQuant 
     & \cellcolor[rgb]{0.843, 0.980, 0.816}1024 
     & \cellcolor[rgb]{0.843, 0.980, 0.816}4/4 
     & \cellcolor[rgb]{0.843, 0.980, 0.816}\textbf{8.01} 
     & \cellcolor[rgb]{0.843, 0.980, 0.816}\textbf{13.92} 
     & \cellcolor[rgb]{0.843, 0.980, 0.816}\textbf{257.24} \\
    \bottomrule
\end{tabular}
    }
    \vspace{-0.3cm}
\end{table}
\begin{table}[!t]\footnotesize 
    \centering
    \caption{Results of text-conditional image generation.}
    \vspace{-0.2cm}
    \label{tab:con1}
    \setlength{\tabcolsep}{1.0mm}
    \resizebox{\linewidth}{!}{
    \begin{tabular}{ccccccc}
    \toprule
    Task & Method & Calib. & Prec. & FID~$\downarrow$ & sFID~$\downarrow$ & CLIP~$\uparrow$ \\
    \cmidrule(r){1-7}
    \multirow{9}{1.7cm}{\centering MS-COCO \\512 $\times $ 512\\$ $ \\Stable-Diffusion\\steps = 50\\eta=0.0\\scale = 7.5} & FP & - & 32/32 & 21.96 & 33.86 & 26.88 \\ 
    \cmidrule(r){2-7}
     & EDA-DM~\textsuperscript{$\dagger$} & 512 & 6/6 & N/A & N/A & N/A \\
     & TFMQ-DM~\textsuperscript{$\dagger$} & 1024 & 6/6 & 165.21 & 124.80 & 18.58 \\
     & EfficientDM~\textsuperscript{$\star$} & 12.8K & 6/6 & 154.61 & 74.50 & 19.01 \\
     & \cellcolor[rgb]{0.843, 0.980, 0.816}DilateQuant 
     & \cellcolor[rgb]{0.843, 0.980, 0.816}512 
     & \cellcolor[rgb]{0.843, 0.980, 0.816}6/6 
     & \cellcolor[rgb]{0.843, 0.980, 0.816}\textbf{24.69} 
     & \cellcolor[rgb]{0.843, 0.980, 0.816}\textbf{33.06} 
     & \cellcolor[rgb]{0.843, 0.980, 0.816}\textbf{26.62} \\
    \cmidrule(r){2-7}
     & EDA-DM~\textsuperscript{$\dagger$} & 512 & 4/4 & N/A & N/A & N/A \\
     & TFMQ-DM~\textsuperscript{$\dagger$} & 1024 & 4/4 & 459.33 & 313.92 & 13.07 \\
     & EfficientDM~\textsuperscript{$\star$} & 12.8K & 4/4 & 216.43 & 111.76 & 14.35 \\
     & \cellcolor[rgb]{0.843, 0.980, 0.816}DilateQuant 
     & \cellcolor[rgb]{0.843, 0.980, 0.816}512 
     & \cellcolor[rgb]{0.843, 0.980, 0.816}4/4 
     & \cellcolor[rgb]{0.843, 0.980, 0.816}\textbf{44.82} 
     & \cellcolor[rgb]{0.843, 0.980, 0.816}\textbf{42.97} 
     & \cellcolor[rgb]{0.843, 0.980, 0.816}\textbf{23.51} \\
    \bottomrule
\end{tabular}
    }
    \vspace{-0.3cm}
\end{table}
\begin{table}[!t]\footnotesize
    \centering
    \caption{Results of DiT model generation. Here, ``Time" and ``Memory" represent the time cost and the peak GPU memory consumption during the quantization process, respectively.}
    \vspace{-0.2cm}
    \label{tab:dit}
    \setlength{\tabcolsep}{1.0mm}
    \resizebox{\linewidth}{!}{
    \begin{tabular}{cc | ccc | ccc}
    \toprule
    \multirow{2}{0.8cm}{\centering Method}& \multirow{2}{0.6cm}{\centering Prec.} & \multicolumn{3}{c|}{\centering \bf Accuracy} & \multicolumn{3}{c}{\centering \bf Efficiency}  \\
     & & FID~$\downarrow$ & sFID~$\downarrow$ & IS~$\uparrow$ & {\centering Calib.} & {\centering Time} & {\centering Memory} \\
    \cmidrule(r){1-8}
    PTQ4DiT~\textsuperscript{$\dagger$} & 6/6 & 20.68 & 42.56 & 103.24 & 8000 & 14.50 h & 15564 MB \\
    \rowcolor[rgb]{0.843, 0.980, 0.816}Ours & 6/6 & \textbf{15.63} & \textbf{31.58} & \textbf{157.64} & 5120 & 4.63 h & 15686 MB \\
    \cmidrule(r){1-8}
    PTQ4DiT~\textsuperscript{$\dagger$} & 4/4 & 256.80 & 140.54 & 2.27 & 8000 & 14.50 h & 15564 MB \\
    \rowcolor[rgb]{0.843, 0.980, 0.816}Ours & 4/4 & \textbf{56.83} & \textbf{54.57} & \textbf{89.66} & 5120 & 4.63 h & 15686 MB \\
    \bottomrule
\end{tabular}
    }
    \vspace{-0.3cm}
\end{table}
\paragraph{Quantization and Comparison Settings.}
We employ DilateQuant with the standard channel-wise quantization for weights and layer-wise quantization for activations.
To highlight the efficiency, DilateQuant selects 5120 samples for calibration and trains for 5K iterations with a batch size of 32, aligning with PTQ-based method~\citep{liu2024enhanced}.
The Adam~\citep{kingma2014adam} optimizer is adopted, and the learning rates for quantization parameters and weights are set as 1e-4 and 1e-2, respectively.
For the experimental comparison, we compare DilateQuant with PTQ-based method (PTQD~\cite{he2023ptqd}, EDA-DM~\citep{liu2024enhanced}, TFMQ-DM~\cite{huang2024tfmq}, TCAQ-DM~\citep{huang2024tcaq} and PTQ4DiT~\cite{wu2024ptq4dit}) and variant QAT-based methods (EfficientDM~\citep{he2023efficientdm} and QuEST~\citep{wang2024quest}).
Notably, since these two variant QAT-based methods employ non-standard settings, we modify them to follow standard settings for a fair comparison.
As a result, some of the reported results may differ from those in the original papers.
To further compare with them, we also employ the same non-standard settings on DilateQuant to conduct experiments in Appendix E. 
Moreover, considering that SVDQuant~\citep{li2024svdqunat} introduces additional computation during inference and requires customized CUDA kernel support, we exclude it from our comparisons.

\subsection{Main Result}
\paragraph{Unconditional Generation.}
As reported in Table~\ref{tab:uncon}, at 4-bit precision, previous works all suffer from non-trivial performance degradation. For instance, EDA-DM and QuEST become infeasible on LSUN-Bedroom, and EfficientDM remains far from practical usability on LSUN-Church.
In contrast, DilateQuant achieves a substantial improvement in performance, with encouraging 6.28 and 4.98 FID improvement over EfficientDM on two LSUN datasets, respectively.
Additionally, at 6-bit precision, DilateQuant achieves a fidelity comparable to that of the full-precision (FP) baseline.
\paragraph{Conditional Generation.}
The results for conditional generation are reported in Table~\ref{tab:con} and~\ref{tab:con1}.
When the bit-width is reduced to 4-bit, PTQ-based methods struggle in class-conditional generation tasks.
Fortunately, DilateQuant preserves competitive performance, achieving a 4.07 improvement in FID (8.01 vs. 12.08) and a 135.12 gain in IS (257.24 vs. 122.12) compared to EfficientDM.
For text-conditional generation, DilateQuant reduces the FID of Stable Diffusion to 24.69 at 6-bit precision and maintains usable performance even at 4-bit precision.
\paragraph{Generation of DiT Models.}
We further extend DilateQuant to the DiT-XL/2 models on ImageNet (256 $\times $ 256). Following PTQ4DiT~\cite{wu2024ptq4dit}, we evaluate using the DiT-XL/2 model with 50 steps.
As shown in Table~\ref{tab:dit}, our method significantly outperforms PTQ4DiT in both accuracy and efficiency. 
Specifically, compared to PTQ4DiT, DilateQuant improves performance across various accuracy metrics. Additionally, DilateQuant requires only 4.63 hours of training compared to the 14.50 hours needed for PTQ4DiT.
\begin{table}[!t]\small 
    \centering
    \caption{Aesthetic assessment at 4-bit precision.}
    \vspace{-0.2cm}
    \label{tab:aes}
    \setlength{\tabcolsep}{3.0mm}
    \resizebox{\linewidth}{!}{
    \begin{tabular}{cccc}
    \toprule
    Method & LSUN-Bedroom & ImageNet & DrawBench \\
    \cmidrule(r){1-4}
    FP & 5.91 & 5.32 & 5.80 \\
    \cmidrule(r){1-4}
    EfficientDM & 5.47 & 3.51 & 2.84 \\
    \rowcolor[rgb]{0.843, 0.980, 0.816}DilateQuant & \textbf{5.72} & \textbf{4.85} & \textbf{5.23} \\
    \bottomrule
\end{tabular}
    }
    \vspace{-0.3cm}
\end{table}

\begin{figure*}[!t]
    \centering
    \includegraphics[width=1.0\textwidth]{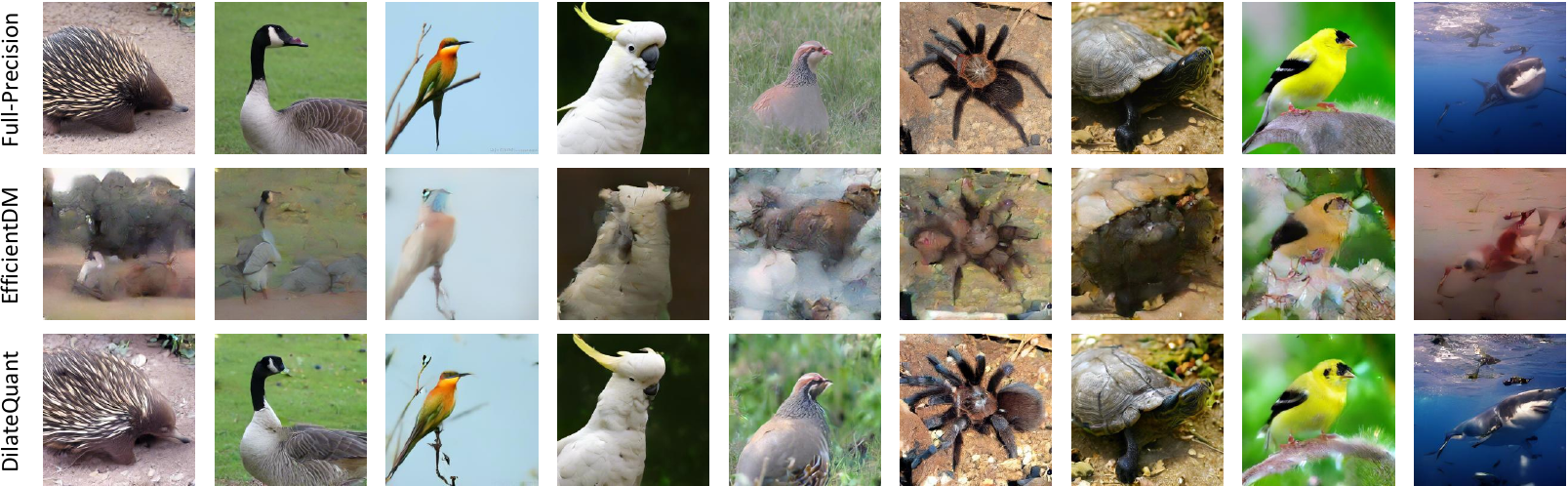}
    \vspace{-0.5cm}
    \caption{Random samples of quantized models on ImageNet with 4-bit quantization.}
    \label{fig:rand}
    \vspace{-0.2cm}
\end{figure*}

\begin{figure}[!t]
    \centering
    \includegraphics[width=0.92\linewidth]{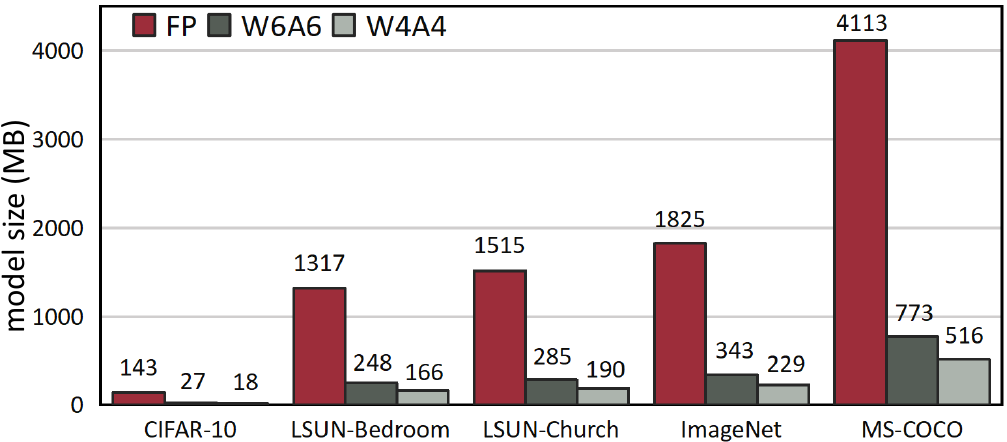}
    \vspace{-0.2cm}
    \caption{Model sizes of quantized diffusion models.}
    \label{fig:compression}
    \vspace{-0.4cm}
\end{figure}

\paragraph{Human Preference Evaluation.}
Since automated metrics do not fully reflect the quality of generation, we use \emph{Aesthetic Predictor} to evaluate Aesthetic Score~$\uparrow$, mimicking human preference assessment of the generated images.
For Stable-Diffusion, we use the convincing DrawBench~\cite{saharia2022photorealistic} benchmark to evaluate.
As reported in Table~\ref{tab:aes} and Figure~\ref{fig:rand}, DilateQuant has a better aesthetic representation compared to EfficientDM.
We also visualize the random samples of quantization results in Appendix J. 
\paragraph{Quantized Model Deployment.}
To visualize the acceleration and compression effects of quantization, we deploy the quantized model on an RTX3090 GPU.
As reported in Table~\ref{tab:deploy}, the bit operations (Bops) of network are reduced from 102.3 T to 1.7 T after quantization, while the runtime required to generate an image decreases from 436.8 ms to 130.4 ms, achieving a 3.35$\times$ speedup.
The compression effects of models at different bit-widths are shown in Figure~\ref{fig:compression}. DilateQuant significantly reduces the model size of Stable-Diffusion from 4113 MB to 516 MB, effectively advancing the practical applications of it in real-world scenarios.

\begin{table}[!t]\small 
    \centering
    \caption{Real-world evaluation of LDM-4 on ImageNet.}
    \vspace{-0.2cm}
    \setlength{\tabcolsep}{1.3mm}
    \begin{tabular}{c|ccccc}
        \toprule
        Method & Prec. & Bops & Model Size & Runtime & Speedup \\
        \midrule
        LDM-4 & 32/32 & 102.3 T & 1824.6 MB & 436.8 ms & 1.00$\times$ \\
        Ours & 4/4 & 1.7 T & 229.2 MB & 130.4 ms & 3.35$\times$ \\
        \bottomrule
    \end{tabular}
    \label{tab:deploy}
    \vspace{-0.3cm}
\end{table}

\subsection{Ablation Study}
The ablation experiments are conducted over DDIM on CIFAR-10 with 4-bit quantization.
We start by analysing the efficacy of each proposed component, as reported in Table~\ref{tab:ablation1}.
Our study adopts the PTQ-based EDA-DM~\citep{liu2024enhanced} as the baseline, aiming to enhance model performance while maintaining quantization efficiency.
By incorporating WD, which effectively alleviates the wide activation range, we achieve a significant improvement in FID, reaching 26.26. Further integration of TPQ, our approachs push the performance limits of PTQ-based methods to achieve an FID score of 16.27.
The introduction of BKD transforms the approach into a QAT framework, as it involves retraining the quantized weight of models.
By combining BKD, DilateQuant reduces the FID score to 9.13, achieving a generation quality comparable to that of full-precision models.

We also conduct the efficiency analysis of DilateQuant by comparing it with PTQ~\citep{liu2024enhanced}, QAT~\citep{esser2019learned}, and variants of QAT (V-QAT)~\citep{he2023efficientdm,wang2024quest} methods.
As reported in Table~\ref{tab:ablation2}, the PTQ method fails to maintain performance and the QAT method requires significant resources.
In sharp contrast, DilateQuant achieves QAT-like accuracy with PTQ-like time cost and GPU memory.

The efficiency comparisons on other models are reported in Appendix D. 
We also add the ablation experiments of DilateQuant for time steps and samplers in Appendix C. 
\begin{table}[t]\footnotesize 
    \centering
    \caption{Efficacy of different component in this paper.}
    \vspace{-0.2cm}
    \label{tab:ablation1}
    \setlength{\tabcolsep}{1.1mm}
    \resizebox{\linewidth}{!}{
    \begin{tabular}{ccccccccc}
    \toprule 
    \multicolumn{3}{c}{Method} & \multirow{2}{1.0cm}{\centering Framework} & \multirow{2}{0.55cm}{\centering Prec.} & \multirow{2}{0.55cm}{\centering FID~$\downarrow$} & \multirow{2}{0.55cm}{\centering sFID~$\downarrow$} & \multirow{2}{0.55cm}{\centering IS~$\uparrow$} & \multirow{2}{0.7cm}{\centering Time(h)} \\ 
    WD & TPQ & BKD & & & & & & \\ 
    \midrule
    \xmark & \xmark & \xmark & PTQ & 4/4 & 120.24 & 36.72 & 4.42 & 0.97 \\ 
    \cmark & \xmark & \xmark & PTQ & 4/4 & 26.26 & 16.73 & 7.78 & 0.97 \\ 
    \cmark & \cmark & \xmark & PTQ & 4/4 & 16.27 & 11.83 & 8.09 & 0.98 \\ 
    \cmark & \cmark & \cmark & V-QAT & 4/4 & 9.13 & 6.92 & 8.56 & 1.08 \\ 
    \bottomrule
\end{tabular}
    }
    \vspace{-0.3cm}
\end{table}
\begin{table}[t]\footnotesize 
    \centering
    \caption{Efficiency comparisons of various quantization frameworks. Here, ``Data" denotes the number of original datasets used.}
    \vspace{-0.2cm}
    \label{tab:ablation2}
    \setlength{\tabcolsep}{0.8mm}
    \resizebox{\linewidth}{!}{
    \begin{tabular}{ccccccc}
    \toprule 
    Task & Method & Framework & Calib. & Data & Time & Memory \\ 
    \cmidrule(r){1-7}
    \multirow{4}{1.5cm}{\centering CIFAR-10 \\32 $\times $ 32} & EDA-DM & PTQ & 5120 & 0 & 0.97 h & 3019 MB \\ 
    \multirow{4}{*}{} & LSQ & QAT & - & 50K & 13.89 h & 9974 MB \\ 
    \multirow{4}{*}{} & EfficientDM & V-QAT & 819.2K & 0 & 2.98 h & 9546 MB \\ 
    \multirow{4}{*}{} & Ours & V-QAT & 5120 & 0 & 1.08 h & 3439 MB \\ 
    \cmidrule(r){1-7}
    \multirow{2}{1.5cm}{\centering ImageNet \\256 $\times $ 256} & QuEST & V-QAT & 5120 & 0 & 15.25 h & 20642 MB \\ 
    \multirow{2}{*}{} & Ours & V-QAT & 1024 & 0 & 6.56 h & 14680 MB \\ 
    \bottomrule
\end{tabular}
    }
    \vspace{-0.4cm}
\end{table}

\section{Conclusion}
In this work, we propose DilateQuant, a novel QAT framework for diffusion models that offers comparable accuracy and high efficiency.
Specifically, we find the unsaturation property of the in-channel weights and exploit it to alleviate the wide range of activations.
By dilating the unsaturated channels to a constrained range, our method steadily minimizes quantization errors and ensures the convergence of QAT training.
Furthermore, we design a flexible quantizer and introduce a novel knowledge distillation strategy to further enhance performance while significantly improving training efficiency. 
Extensive experiments demonstrate that DilateQuant significantly outperforms existing methods in low-bit quantization.
More importantly, it provides valuable insights for designing efficient QAT frameworks.

\newpage
\bibliographystyle{ACM-Reference-Format}
\bibliography{sample-base}


\begin{thebibliography}{67}


\ifx \showCODEN    \undefined \def \showCODEN     #1{\unskip}     \fi
\ifx \showDOI      \undefined \def \showDOI       #1{#1}\fi
\ifx \showISBNx    \undefined \def \showISBNx     #1{\unskip}     \fi
\ifx \showISBNxiii \undefined \def \showISBNxiii  #1{\unskip}     \fi
\ifx \showISSN     \undefined \def \showISSN      #1{\unskip}     \fi
\ifx \showLCCN     \undefined \def \showLCCN      #1{\unskip}     \fi
\ifx \shownote     \undefined \def \shownote      #1{#1}          \fi
\ifx \showarticletitle \undefined \def \showarticletitle #1{#1}   \fi
\ifx \showURL      \undefined \def \showURL       {\relax}        \fi
\providecommand\bibfield[2]{#2}
\providecommand\bibinfo[2]{#2}
\providecommand\natexlab[1]{#1}
\providecommand\showeprint[2][]{arXiv:#2}

\bibitem[Chen et~al\mbox{.}(2024)]%
        {chen2024delta}
\bibfield{author}{\bibinfo{person}{Pengtao Chen}, \bibinfo{person}{Mingzhu Shen}, \bibinfo{person}{Peng Ye}, \bibinfo{person}{Jianjian Cao}, \bibinfo{person}{Chongjun Tu}, \bibinfo{person}{Christos-Savvas Bouganis}, \bibinfo{person}{Yiren Zhao}, {and} \bibinfo{person}{Tao Chen}.} \bibinfo{year}{2024}\natexlab{}.
\newblock \showarticletitle{Delta-DiT: A Training-Free Acceleration Method Tailored for Diffusion Transformers}.
\newblock \bibinfo{journal}{\emph{arXiv preprint arXiv:2406.01125}} (\bibinfo{year}{2024}).
\newblock


\bibitem[Deng et~al\mbox{.}(2009)]%
        {deng2009imagenet}
\bibfield{author}{\bibinfo{person}{Jia Deng}, \bibinfo{person}{Wei Dong}, \bibinfo{person}{Richard Socher}, \bibinfo{person}{Li-Jia Li}, \bibinfo{person}{Kai Li}, {and} \bibinfo{person}{Li Fei-Fei}.} \bibinfo{year}{2009}\natexlab{}.
\newblock \showarticletitle{Imagenet: A large-scale hierarchical image database}. In \bibinfo{booktitle}{\emph{2009 IEEE conference on computer vision and pattern recognition}}. Ieee, \bibinfo{pages}{248--255}.
\newblock


\bibitem[Ding et~al\mbox{.}(2023)]%
        {ding2023cbq}
\bibfield{author}{\bibinfo{person}{Xin Ding}, \bibinfo{person}{Xiaoyu Liu}, \bibinfo{person}{Zhijun Tu}, \bibinfo{person}{Yun Zhang}, \bibinfo{person}{Wei Li}, \bibinfo{person}{Jie Hu}, \bibinfo{person}{Hanting Chen}, \bibinfo{person}{Yehui Tang}, \bibinfo{person}{Zhiwei Xiong}, \bibinfo{person}{Baoqun Yin}, {et~al\mbox{.}}} \bibinfo{year}{2023}\natexlab{}.
\newblock \showarticletitle{Cbq: Cross-block quantization for large language models}.
\newblock \bibinfo{journal}{\emph{arXiv preprint arXiv:2312.07950}} (\bibinfo{year}{2023}).
\newblock


\bibitem[Esser et~al\mbox{.}(2019)]%
        {esser2019learned}
\bibfield{author}{\bibinfo{person}{Steven~K Esser}, \bibinfo{person}{Jeffrey~L McKinstry}, \bibinfo{person}{Deepika Bablani}, \bibinfo{person}{Rathinakumar Appuswamy}, {and} \bibinfo{person}{Dharmendra~S Modha}.} \bibinfo{year}{2019}\natexlab{}.
\newblock \showarticletitle{Learned step size quantization}.
\newblock \bibinfo{journal}{\emph{arXiv preprint arXiv:1902.08153}} (\bibinfo{year}{2019}).
\newblock


\bibitem[Gong et~al\mbox{.}(2019)]%
        {gong2019differentiable}
\bibfield{author}{\bibinfo{person}{Ruihao Gong}, \bibinfo{person}{Xianglong Liu}, \bibinfo{person}{Shenghu Jiang}, \bibinfo{person}{Tianxiang Li}, \bibinfo{person}{Peng Hu}, \bibinfo{person}{Jiazhen Lin}, \bibinfo{person}{Fengwei Yu}, {and} \bibinfo{person}{Junjie Yan}.} \bibinfo{year}{2019}\natexlab{}.
\newblock \showarticletitle{Differentiable soft quantization: Bridging full-precision and low-bit neural networks}. In \bibinfo{booktitle}{\emph{Proceedings of the IEEE/CVF international conference on computer vision}}. \bibinfo{pages}{4852--4861}.
\newblock


\bibitem[He et~al\mbox{.}(2023b)]%
        {he2023efficientdm}
\bibfield{author}{\bibinfo{person}{Yefei He}, \bibinfo{person}{Jing Liu}, \bibinfo{person}{Weijia Wu}, \bibinfo{person}{Hong Zhou}, {and} \bibinfo{person}{Bohan Zhuang}.} \bibinfo{year}{2023}\natexlab{b}.
\newblock \showarticletitle{Efficientdm: Efficient quantization-aware fine-tuning of low-bit diffusion models}.
\newblock \bibinfo{journal}{\emph{arXiv preprint arXiv:2310.03270}} (\bibinfo{year}{2023}).
\newblock


\bibitem[He et~al\mbox{.}(2023a)]%
        {he2023ptqd}
\bibfield{author}{\bibinfo{person}{Yefei He}, \bibinfo{person}{Luping Liu}, \bibinfo{person}{Jing Liu}, \bibinfo{person}{Weijia Wu}, \bibinfo{person}{Hong Zhou}, {and} \bibinfo{person}{Bohan Zhuang}.} \bibinfo{year}{2023}\natexlab{a}.
\newblock \showarticletitle{PTQD: Accurate Post-Training Quantization for Diffusion Models}.
\newblock \bibinfo{journal}{\emph{arXiv preprint arXiv:2305.10657}} (\bibinfo{year}{2023}).
\newblock


\bibitem[Hessel et~al\mbox{.}(2021)]%
        {hessel2021clipscore}
\bibfield{author}{\bibinfo{person}{Jack Hessel}, \bibinfo{person}{Ari Holtzman}, \bibinfo{person}{Maxwell Forbes}, \bibinfo{person}{Ronan~Le Bras}, {and} \bibinfo{person}{Yejin Choi}.} \bibinfo{year}{2021}\natexlab{}.
\newblock \showarticletitle{Clipscore: A reference-free evaluation metric for image captioning}.
\newblock \bibinfo{journal}{\emph{arXiv preprint arXiv:2104.08718}} (\bibinfo{year}{2021}).
\newblock


\bibitem[Heusel et~al\mbox{.}(2017)]%
        {heusel2017gans}
\bibfield{author}{\bibinfo{person}{Martin Heusel}, \bibinfo{person}{Hubert Ramsauer}, \bibinfo{person}{Thomas Unterthiner}, \bibinfo{person}{Bernhard Nessler}, {and} \bibinfo{person}{Sepp Hochreiter}.} \bibinfo{year}{2017}\natexlab{}.
\newblock \showarticletitle{Gans trained by a two time-scale update rule converge to a local nash equilibrium}.
\newblock \bibinfo{journal}{\emph{Advances in neural information processing systems}}  \bibinfo{volume}{30} (\bibinfo{year}{2017}).
\newblock


\bibitem[Ho et~al\mbox{.}(2022)]%
        {ho2022imagen}
\bibfield{author}{\bibinfo{person}{Jonathan Ho}, \bibinfo{person}{William Chan}, \bibinfo{person}{Chitwan Saharia}, \bibinfo{person}{Jay Whang}, \bibinfo{person}{Ruiqi Gao}, \bibinfo{person}{Alexey Gritsenko}, \bibinfo{person}{Diederik~P Kingma}, \bibinfo{person}{Ben Poole}, \bibinfo{person}{Mohammad Norouzi}, \bibinfo{person}{David~J Fleet}, {et~al\mbox{.}}} \bibinfo{year}{2022}\natexlab{}.
\newblock \showarticletitle{Imagen video: High definition video generation with diffusion models}.
\newblock \bibinfo{journal}{\emph{arXiv preprint arXiv:2210.02303}} (\bibinfo{year}{2022}).
\newblock


\bibitem[Ho et~al\mbox{.}(2020)]%
        {ho2020denoising}
\bibfield{author}{\bibinfo{person}{Jonathan Ho}, \bibinfo{person}{Ajay Jain}, {and} \bibinfo{person}{Pieter Abbeel}.} \bibinfo{year}{2020}\natexlab{}.
\newblock \showarticletitle{Denoising diffusion probabilistic models}.
\newblock \bibinfo{journal}{\emph{Advances in neural information processing systems}}  \bibinfo{volume}{33} (\bibinfo{year}{2020}), \bibinfo{pages}{6840--6851}.
\newblock


\bibitem[Hoedt and Klambauer(2024)]%
        {hoedt2024principled}
\bibfield{author}{\bibinfo{person}{Pieter-Jan Hoedt} {and} \bibinfo{person}{G{\"u}nter Klambauer}.} \bibinfo{year}{2024}\natexlab{}.
\newblock \showarticletitle{Principled weight initialisation for input-convex neural networks}.
\newblock \bibinfo{journal}{\emph{Advances in Neural Information Processing Systems}}  \bibinfo{volume}{36} (\bibinfo{year}{2024}).
\newblock


\bibitem[Huang et~al\mbox{.}(2024a)]%
        {huang2024tcaq}
\bibfield{author}{\bibinfo{person}{Haocheng Huang}, \bibinfo{person}{Jiaxin Chen}, \bibinfo{person}{Jinyang Guo}, \bibinfo{person}{Ruiyi Zhan}, {and} \bibinfo{person}{Yunhong Wang}.} \bibinfo{year}{2024}\natexlab{a}.
\newblock \showarticletitle{TCAQ-DM: Timestep-Channel Adaptive Quantization for Diffusion Models}.
\newblock \bibinfo{journal}{\emph{arXiv preprint arXiv:2412.16700}} (\bibinfo{year}{2024}).
\newblock


\bibitem[Huang et~al\mbox{.}(2024b)]%
        {huang2024tfmq}
\bibfield{author}{\bibinfo{person}{Yushi Huang}, \bibinfo{person}{Ruihao Gong}, \bibinfo{person}{Jing Liu}, \bibinfo{person}{Tianlong Chen}, {and} \bibinfo{person}{Xianglong Liu}.} \bibinfo{year}{2024}\natexlab{b}.
\newblock \showarticletitle{Tfmq-dm: Temporal feature maintenance quantization for diffusion models}. In \bibinfo{booktitle}{\emph{Proceedings of the IEEE/CVF Conference on Computer Vision and Pattern Recognition}}. \bibinfo{pages}{7362--7371}.
\newblock


\bibitem[Jacob et~al\mbox{.}(2018)]%
        {jacob2018quantization}
\bibfield{author}{\bibinfo{person}{Benoit Jacob}, \bibinfo{person}{Skirmantas Kligys}, \bibinfo{person}{Bo Chen}, \bibinfo{person}{Menglong Zhu}, \bibinfo{person}{Matthew Tang}, \bibinfo{person}{Andrew Howard}, \bibinfo{person}{Hartwig Adam}, {and} \bibinfo{person}{Dmitry Kalenichenko}.} \bibinfo{year}{2018}\natexlab{}.
\newblock \showarticletitle{Quantization and training of neural networks for efficient integer-arithmetic-only inference}. In \bibinfo{booktitle}{\emph{Proceedings of the IEEE conference on computer vision and pattern recognition}}. \bibinfo{pages}{2704--2713}.
\newblock


\bibitem[Kim et~al\mbox{.}(2022)]%
        {kim2022integer}
\bibfield{author}{\bibinfo{person}{Sehoon Kim}, \bibinfo{person}{Amir Gholami}, \bibinfo{person}{Zhewei Yao}, \bibinfo{person}{Nicholas Lee}, \bibinfo{person}{Patrick Wang}, \bibinfo{person}{Aniruddha Nrusimha}, \bibinfo{person}{Bohan Zhai}, \bibinfo{person}{Tianren Gao}, \bibinfo{person}{Michael~W Mahoney}, {and} \bibinfo{person}{Kurt Keutzer}.} \bibinfo{year}{2022}\natexlab{}.
\newblock \showarticletitle{Integer-only zero-shot quantization for efficient speech recognition}. In \bibinfo{booktitle}{\emph{ICASSP 2022-2022 IEEE International Conference on Acoustics, Speech and Signal Processing (ICASSP)}}. IEEE, \bibinfo{pages}{4288--4292}.
\newblock


\bibitem[Kingma and Ba(2014)]%
        {kingma2014adam}
\bibfield{author}{\bibinfo{person}{Diederik~P Kingma} {and} \bibinfo{person}{Jimmy Ba}.} \bibinfo{year}{2014}\natexlab{}.
\newblock \showarticletitle{Adam: A method for stochastic optimization}.
\newblock \bibinfo{journal}{\emph{arXiv preprint arXiv:1412.6980}} (\bibinfo{year}{2014}).
\newblock


\bibitem[Krizhevsky et~al\mbox{.}(2009)]%
        {krizhevsky2009learning}
\bibfield{author}{\bibinfo{person}{Alex Krizhevsky}, \bibinfo{person}{Geoffrey Hinton}, {et~al\mbox{.}}} \bibinfo{year}{2009}\natexlab{}.
\newblock \showarticletitle{Learning multiple layers of features from tiny images}.
\newblock  (\bibinfo{year}{2009}).
\newblock


\bibitem[Li et~al\mbox{.}(2022)]%
        {li2022srdiff}
\bibfield{author}{\bibinfo{person}{Haoying Li}, \bibinfo{person}{Yifan Yang}, \bibinfo{person}{Meng Chang}, \bibinfo{person}{Shiqi Chen}, \bibinfo{person}{Huajun Feng}, \bibinfo{person}{Zhihai Xu}, \bibinfo{person}{Qi Li}, {and} \bibinfo{person}{Yueting Chen}.} \bibinfo{year}{2022}\natexlab{}.
\newblock \showarticletitle{Srdiff: Single image super-resolution with diffusion probabilistic models}.
\newblock \bibinfo{journal}{\emph{Neurocomputing}}  \bibinfo{volume}{479} (\bibinfo{year}{2022}), \bibinfo{pages}{47--59}.
\newblock


\bibitem[Li et~al\mbox{.}(2024a)]%
        {li2024svdqunat}
\bibfield{author}{\bibinfo{person}{Muyang Li}, \bibinfo{person}{Yujun Lin}, \bibinfo{person}{Zhekai Zhang}, \bibinfo{person}{Tianle Cai}, \bibinfo{person}{Xiuyu Li}, \bibinfo{person}{Junxian Guo}, \bibinfo{person}{Enze Xie}, \bibinfo{person}{Chenlin Meng}, \bibinfo{person}{Jun-Yan Zhu}, {and} \bibinfo{person}{Song Han}.} \bibinfo{year}{2024}\natexlab{a}.
\newblock \showarticletitle{Svdqunat: Absorbing outliers by low-rank components for 4-bit diffusion models}.
\newblock \bibinfo{journal}{\emph{arXiv preprint arXiv:2411.05007}} (\bibinfo{year}{2024}).
\newblock


\bibitem[Li et~al\mbox{.}(2023a)]%
        {li2023q}
\bibfield{author}{\bibinfo{person}{Xiuyu Li}, \bibinfo{person}{Yijiang Liu}, \bibinfo{person}{Long Lian}, \bibinfo{person}{Huanrui Yang}, \bibinfo{person}{Zhen Dong}, \bibinfo{person}{Daniel Kang}, \bibinfo{person}{Shanghang Zhang}, {and} \bibinfo{person}{Kurt Keutzer}.} \bibinfo{year}{2023}\natexlab{a}.
\newblock \showarticletitle{Q-diffusion: Quantizing diffusion models}. In \bibinfo{booktitle}{\emph{Proceedings of the IEEE/CVF International Conference on Computer Vision}}. \bibinfo{pages}{17535--17545}.
\newblock


\bibitem[Li et~al\mbox{.}(2021)]%
        {li2021brecq}
\bibfield{author}{\bibinfo{person}{Yuhang Li}, \bibinfo{person}{Ruihao Gong}, \bibinfo{person}{Xu Tan}, \bibinfo{person}{Yang Yang}, \bibinfo{person}{Peng Hu}, \bibinfo{person}{Qi Zhang}, \bibinfo{person}{Fengwei Yu}, \bibinfo{person}{Wei Wang}, {and} \bibinfo{person}{Shi Gu}.} \bibinfo{year}{2021}\natexlab{}.
\newblock \showarticletitle{Brecq: Pushing the limit of post-training quantization by block reconstruction}.
\newblock \bibinfo{journal}{\emph{arXiv preprint arXiv:2102.05426}} (\bibinfo{year}{2021}).
\newblock


\bibitem[Li and Gu(2023)]%
        {li2023vit}
\bibfield{author}{\bibinfo{person}{Zhikai Li} {and} \bibinfo{person}{Qingyi Gu}.} \bibinfo{year}{2023}\natexlab{}.
\newblock \showarticletitle{I-vit: Integer-only quantization for efficient vision transformer inference}. In \bibinfo{booktitle}{\emph{Proceedings of the IEEE/CVF International Conference on Computer Vision}}. \bibinfo{pages}{17065--17075}.
\newblock


\bibitem[Li et~al\mbox{.}(2025)]%
        {li2025k}
\bibfield{author}{\bibinfo{person}{Zhikai Li}, \bibinfo{person}{Xuewen Liu}, \bibinfo{person}{Dongrong~Joe Fu}, \bibinfo{person}{Jianquan Li}, \bibinfo{person}{Qingyi Gu}, \bibinfo{person}{Kurt Keutzer}, {and} \bibinfo{person}{Zhen Dong}.} \bibinfo{year}{2025}\natexlab{}.
\newblock \showarticletitle{K-sort arena: Efficient and reliable benchmarking for generative models via k-wise human preferences}. In \bibinfo{booktitle}{\emph{Proceedings of the Computer Vision and Pattern Recognition Conference}}. \bibinfo{pages}{9131--9141}.
\newblock


\bibitem[Li et~al\mbox{.}(2024b)]%
        {li2024repquant}
\bibfield{author}{\bibinfo{person}{Zhikai Li}, \bibinfo{person}{Xuewen Liu}, \bibinfo{person}{Jing Zhang}, {and} \bibinfo{person}{Qingyi Gu}.} \bibinfo{year}{2024}\natexlab{b}.
\newblock \showarticletitle{Repquant: Towards accurate post-training quantization of large transformer models via scale reparameterization}.
\newblock \bibinfo{journal}{\emph{arXiv preprint arXiv:2402.05628}} (\bibinfo{year}{2024}).
\newblock


\bibitem[Li et~al\mbox{.}(2023b)]%
        {li2023repq}
\bibfield{author}{\bibinfo{person}{Zhikai Li}, \bibinfo{person}{Junrui Xiao}, \bibinfo{person}{Lianwei Yang}, {and} \bibinfo{person}{Qingyi Gu}.} \bibinfo{year}{2023}\natexlab{b}.
\newblock \showarticletitle{Repq-vit: Scale reparameterization for post-training quantization of vision transformers}. In \bibinfo{booktitle}{\emph{Proceedings of the IEEE/CVF International Conference on Computer Vision}}. \bibinfo{pages}{17227--17236}.
\newblock


\bibitem[Lin et~al\mbox{.}(2024)]%
        {lin2024awq}
\bibfield{author}{\bibinfo{person}{Ji Lin}, \bibinfo{person}{Jiaming Tang}, \bibinfo{person}{Haotian Tang}, \bibinfo{person}{Shang Yang}, \bibinfo{person}{Wei-Ming Chen}, \bibinfo{person}{Wei-Chen Wang}, \bibinfo{person}{Guangxuan Xiao}, \bibinfo{person}{Xingyu Dang}, \bibinfo{person}{Chuang Gan}, {and} \bibinfo{person}{Song Han}.} \bibinfo{year}{2024}\natexlab{}.
\newblock \showarticletitle{AWQ: Activation-aware Weight Quantization for On-Device LLM Compression and Acceleration}.
\newblock \bibinfo{journal}{\emph{Proceedings of Machine Learning and Systems}}  \bibinfo{volume}{6} (\bibinfo{year}{2024}), \bibinfo{pages}{87--100}.
\newblock


\bibitem[Lin et~al\mbox{.}(2014)]%
        {lin2014microsoft}
\bibfield{author}{\bibinfo{person}{Tsung-Yi Lin}, \bibinfo{person}{Michael Maire}, \bibinfo{person}{Serge Belongie}, \bibinfo{person}{James Hays}, \bibinfo{person}{Pietro Perona}, \bibinfo{person}{Deva Ramanan}, \bibinfo{person}{Piotr Doll{\'a}r}, {and} \bibinfo{person}{C~Lawrence Zitnick}.} \bibinfo{year}{2014}\natexlab{}.
\newblock \showarticletitle{Microsoft coco: Common objects in context}. In \bibinfo{booktitle}{\emph{Computer Vision--ECCV 2014: 13th European Conference, Zurich, Switzerland, September 6-12, 2014, Proceedings, Part V 13}}. Springer, \bibinfo{pages}{740--755}.
\newblock


\bibitem[Liu et~al\mbox{.}(2022)]%
        {liu2022pseudo}
\bibfield{author}{\bibinfo{person}{Luping Liu}, \bibinfo{person}{Yi Ren}, \bibinfo{person}{Zhijie Lin}, {and} \bibinfo{person}{Zhou Zhao}.} \bibinfo{year}{2022}\natexlab{}.
\newblock \showarticletitle{Pseudo numerical methods for diffusion models on manifolds}.
\newblock \bibinfo{journal}{\emph{arXiv preprint arXiv:2202.09778}} (\bibinfo{year}{2022}).
\newblock


\bibitem[Liu et~al\mbox{.}(2025)]%
        {liu2025cachequant}
\bibfield{author}{\bibinfo{person}{Xuewen Liu}, \bibinfo{person}{Zhikai Li}, {and} \bibinfo{person}{Qingyi Gu}.} \bibinfo{year}{2025}\natexlab{}.
\newblock \showarticletitle{Cachequant: Comprehensively accelerated diffusion models}. In \bibinfo{booktitle}{\emph{Proceedings of the Computer Vision and Pattern Recognition Conference}}. \bibinfo{pages}{23269--23280}.
\newblock


\bibitem[Liu et~al\mbox{.}(2024)]%
        {liu2024enhanced}
\bibfield{author}{\bibinfo{person}{Xuewen Liu}, \bibinfo{person}{Zhikai Li}, \bibinfo{person}{Junrui Xiao}, {and} \bibinfo{person}{Qingyi Gu}.} \bibinfo{year}{2024}\natexlab{}.
\newblock \showarticletitle{Enhanced distribution alignment for post-training quantization of diffusion models}.
\newblock \bibinfo{journal}{\emph{arXiv preprint arXiv:2401.04585}} (\bibinfo{year}{2024}).
\newblock


\bibitem[Louizos et~al\mbox{.}(2018)]%
        {louizos2018relaxed}
\bibfield{author}{\bibinfo{person}{Christos Louizos}, \bibinfo{person}{Matthias Reisser}, \bibinfo{person}{Tijmen Blankevoort}, \bibinfo{person}{Efstratios Gavves}, {and} \bibinfo{person}{Max Welling}.} \bibinfo{year}{2018}\natexlab{}.
\newblock \showarticletitle{Relaxed quantization for discretized neural networks}.
\newblock \bibinfo{journal}{\emph{arXiv preprint arXiv:1810.01875}} (\bibinfo{year}{2018}).
\newblock


\bibitem[Lu et~al\mbox{.}(2022)]%
        {lu2022dpm}
\bibfield{author}{\bibinfo{person}{Cheng Lu}, \bibinfo{person}{Yuhao Zhou}, \bibinfo{person}{Fan Bao}, \bibinfo{person}{Jianfei Chen}, \bibinfo{person}{Chongxuan Li}, {and} \bibinfo{person}{Jun Zhu}.} \bibinfo{year}{2022}\natexlab{}.
\newblock \showarticletitle{Dpm-solver++: Fast solver for guided sampling of diffusion probabilistic models}.
\newblock \bibinfo{journal}{\emph{arXiv preprint arXiv:2211.01095}} (\bibinfo{year}{2022}).
\newblock


\bibitem[Ma et~al\mbox{.}(2024)]%
        {ma2024deepcache}
\bibfield{author}{\bibinfo{person}{Xinyin Ma}, \bibinfo{person}{Gongfan Fang}, {and} \bibinfo{person}{Xinchao Wang}.} \bibinfo{year}{2024}\natexlab{}.
\newblock \showarticletitle{Deepcache: Accelerating diffusion models for free}. In \bibinfo{booktitle}{\emph{Proceedings of the IEEE/CVF Conference on Computer Vision and Pattern Recognition}}. \bibinfo{pages}{15762--15772}.
\newblock


\bibitem[Nagel et~al\mbox{.}(2020)]%
        {nagel2020up}
\bibfield{author}{\bibinfo{person}{Markus Nagel}, \bibinfo{person}{Rana~Ali Amjad}, \bibinfo{person}{Mart Van~Baalen}, \bibinfo{person}{Christos Louizos}, {and} \bibinfo{person}{Tijmen Blankevoort}.} \bibinfo{year}{2020}\natexlab{}.
\newblock \showarticletitle{Up or down? adaptive rounding for post-training quantization}. In \bibinfo{booktitle}{\emph{International Conference on Machine Learning}}. PMLR, \bibinfo{pages}{7197--7206}.
\newblock


\bibitem[Nagel et~al\mbox{.}(2019)]%
        {nagel2019data}
\bibfield{author}{\bibinfo{person}{Markus Nagel}, \bibinfo{person}{Mart~van Baalen}, \bibinfo{person}{Tijmen Blankevoort}, {and} \bibinfo{person}{Max Welling}.} \bibinfo{year}{2019}\natexlab{}.
\newblock \showarticletitle{Data-free quantization through weight equalization and bias correction}. In \bibinfo{booktitle}{\emph{Proceedings of the IEEE/CVF International Conference on Computer Vision}}. \bibinfo{pages}{1325--1334}.
\newblock


\bibitem[Nichol and Dhariwal(2021)]%
        {nichol2021improved}
\bibfield{author}{\bibinfo{person}{Alexander~Quinn Nichol} {and} \bibinfo{person}{Prafulla Dhariwal}.} \bibinfo{year}{2021}\natexlab{}.
\newblock \showarticletitle{Improved denoising diffusion probabilistic models}. In \bibinfo{booktitle}{\emph{International Conference on Machine Learning}}. PMLR, \bibinfo{pages}{8162--8171}.
\newblock


\bibitem[Niu et~al\mbox{.}(2020)]%
        {niu2020permutation}
\bibfield{author}{\bibinfo{person}{Chenhao Niu}, \bibinfo{person}{Yang Song}, \bibinfo{person}{Jiaming Song}, \bibinfo{person}{Shengjia Zhao}, \bibinfo{person}{Aditya Grover}, {and} \bibinfo{person}{Stefano Ermon}.} \bibinfo{year}{2020}\natexlab{}.
\newblock \showarticletitle{Permutation invariant graph generation via score-based generative modeling}. In \bibinfo{booktitle}{\emph{International Conference on Artificial Intelligence and Statistics}}. PMLR, \bibinfo{pages}{4474--4484}.
\newblock


\bibitem[Peebles and Xie(2023)]%
        {peebles2023scalable}
\bibfield{author}{\bibinfo{person}{William Peebles} {and} \bibinfo{person}{Saining Xie}.} \bibinfo{year}{2023}\natexlab{}.
\newblock \showarticletitle{Scalable diffusion models with transformers}. In \bibinfo{booktitle}{\emph{Proceedings of the IEEE/CVF International Conference on Computer Vision}}. \bibinfo{pages}{4195--4205}.
\newblock


\bibitem[Rombach et~al\mbox{.}(2022)]%
        {rombach2022high}
\bibfield{author}{\bibinfo{person}{Robin Rombach}, \bibinfo{person}{Andreas Blattmann}, \bibinfo{person}{Dominik Lorenz}, \bibinfo{person}{Patrick Esser}, {and} \bibinfo{person}{Bj{\"o}rn Ommer}.} \bibinfo{year}{2022}\natexlab{}.
\newblock \showarticletitle{High-resolution image synthesis with latent diffusion models}. In \bibinfo{booktitle}{\emph{Proceedings of the IEEE/CVF conference on computer vision and pattern recognition}}. \bibinfo{pages}{10684--10695}.
\newblock


\bibitem[Ronneberger et~al\mbox{.}(2015)]%
        {ronneberger2015u}
\bibfield{author}{\bibinfo{person}{Olaf Ronneberger}, \bibinfo{person}{Philipp Fischer}, {and} \bibinfo{person}{Thomas Brox}.} \bibinfo{year}{2015}\natexlab{}.
\newblock \showarticletitle{U-net: Convolutional networks for biomedical image segmentation}. In \bibinfo{booktitle}{\emph{Medical Image Computing and Computer-Assisted Intervention--MICCAI 2015: 18th International Conference, Munich, Germany, October 5-9, 2015, Proceedings, Part III 18}}. Springer, \bibinfo{pages}{234--241}.
\newblock


\bibitem[Saharia et~al\mbox{.}(2022)]%
        {saharia2022photorealistic}
\bibfield{author}{\bibinfo{person}{Chitwan Saharia}, \bibinfo{person}{William Chan}, \bibinfo{person}{Saurabh Saxena}, \bibinfo{person}{Lala Li}, \bibinfo{person}{Jay Whang}, \bibinfo{person}{Emily~L Denton}, \bibinfo{person}{Kamyar Ghasemipour}, \bibinfo{person}{Raphael Gontijo~Lopes}, \bibinfo{person}{Burcu Karagol~Ayan}, \bibinfo{person}{Tim Salimans}, {et~al\mbox{.}}} \bibinfo{year}{2022}\natexlab{}.
\newblock \showarticletitle{Photorealistic text-to-image diffusion models with deep language understanding}.
\newblock \bibinfo{journal}{\emph{Advances in neural information processing systems}}  \bibinfo{volume}{35} (\bibinfo{year}{2022}), \bibinfo{pages}{36479--36494}.
\newblock


\bibitem[Salimans et~al\mbox{.}(2016)]%
        {salimans2016improved}
\bibfield{author}{\bibinfo{person}{Tim Salimans}, \bibinfo{person}{Ian Goodfellow}, \bibinfo{person}{Wojciech Zaremba}, \bibinfo{person}{Vicki Cheung}, \bibinfo{person}{Alec Radford}, {and} \bibinfo{person}{Xi Chen}.} \bibinfo{year}{2016}\natexlab{}.
\newblock \showarticletitle{Improved techniques for training gans}.
\newblock \bibinfo{journal}{\emph{Advances in neural information processing systems}}  \bibinfo{volume}{29} (\bibinfo{year}{2016}).
\newblock


\bibitem[Shang et~al\mbox{.}(2023)]%
        {shang2023post}
\bibfield{author}{\bibinfo{person}{Yuzhang Shang}, \bibinfo{person}{Zhihang Yuan}, \bibinfo{person}{Bin Xie}, \bibinfo{person}{Bingzhe Wu}, {and} \bibinfo{person}{Yan Yan}.} \bibinfo{year}{2023}\natexlab{}.
\newblock \showarticletitle{Post-training quantization on diffusion models}. In \bibinfo{booktitle}{\emph{Proceedings of the IEEE/CVF Conference on Computer Vision and Pattern Recognition}}. \bibinfo{pages}{1972--1981}.
\newblock


\bibitem[Shao et~al\mbox{.}(2023)]%
        {shao2023omniquant}
\bibfield{author}{\bibinfo{person}{Wenqi Shao}, \bibinfo{person}{Mengzhao Chen}, \bibinfo{person}{Zhaoyang Zhang}, \bibinfo{person}{Peng Xu}, \bibinfo{person}{Lirui Zhao}, \bibinfo{person}{Zhiqian Li}, \bibinfo{person}{Kaipeng Zhang}, \bibinfo{person}{Peng Gao}, \bibinfo{person}{Yu Qiao}, {and} \bibinfo{person}{Ping Luo}.} \bibinfo{year}{2023}\natexlab{}.
\newblock \showarticletitle{Omniquant: Omnidirectionally calibrated quantization for large language models}.
\newblock \bibinfo{journal}{\emph{arXiv preprint arXiv:2308.13137}} (\bibinfo{year}{2023}).
\newblock


\bibitem[Shomron et~al\mbox{.}(2021)]%
        {shomron2021post}
\bibfield{author}{\bibinfo{person}{Gil Shomron}, \bibinfo{person}{Freddy Gabbay}, \bibinfo{person}{Samer Kurzum}, {and} \bibinfo{person}{Uri Weiser}.} \bibinfo{year}{2021}\natexlab{}.
\newblock \showarticletitle{Post-training sparsity-aware quantization}.
\newblock \bibinfo{journal}{\emph{Advances in Neural Information Processing Systems}}  \bibinfo{volume}{34} (\bibinfo{year}{2021}), \bibinfo{pages}{17737--17748}.
\newblock


\bibitem[So et~al\mbox{.}(2023)]%
        {so2023temporal}
\bibfield{author}{\bibinfo{person}{Junhyuk So}, \bibinfo{person}{Jungwon Lee}, \bibinfo{person}{Daehyun Ahn}, \bibinfo{person}{Hyungjun Kim}, {and} \bibinfo{person}{Eunhyeok Park}.} \bibinfo{year}{2023}\natexlab{}.
\newblock \showarticletitle{Temporal Dynamic Quantization for Diffusion Models}.
\newblock \bibinfo{journal}{\emph{arXiv preprint arXiv:2306.02316}} (\bibinfo{year}{2023}).
\newblock


\bibitem[Song et~al\mbox{.}(2020)]%
        {song2020denoising}
\bibfield{author}{\bibinfo{person}{Jiaming Song}, \bibinfo{person}{Chenlin Meng}, {and} \bibinfo{person}{Stefano Ermon}.} \bibinfo{year}{2020}\natexlab{}.
\newblock \showarticletitle{Denoising diffusion implicit models}.
\newblock \bibinfo{journal}{\emph{arXiv preprint arXiv:2010.02502}} (\bibinfo{year}{2020}).
\newblock


\bibitem[Song and Ermon(2019)]%
        {song2019generative}
\bibfield{author}{\bibinfo{person}{Yang Song} {and} \bibinfo{person}{Stefano Ermon}.} \bibinfo{year}{2019}\natexlab{}.
\newblock \showarticletitle{Generative modeling by estimating gradients of the data distribution}.
\newblock \bibinfo{journal}{\emph{Advances in neural information processing systems}}  \bibinfo{volume}{32} (\bibinfo{year}{2019}).
\newblock


\bibitem[Sui et~al\mbox{.}(2024)]%
        {sui2024bitsfusion}
\bibfield{author}{\bibinfo{person}{Yang Sui}, \bibinfo{person}{Yanyu Li}, \bibinfo{person}{Anil Kag}, \bibinfo{person}{Yerlan Idelbayev}, \bibinfo{person}{Junli Cao}, \bibinfo{person}{Ju Hu}, \bibinfo{person}{Dhritiman Sagar}, \bibinfo{person}{Bo Yuan}, \bibinfo{person}{Sergey Tulyakov}, {and} \bibinfo{person}{Jian Ren}.} \bibinfo{year}{2024}\natexlab{}.
\newblock \showarticletitle{Bitsfusion: 1.99 bits weight quantization of diffusion model}.
\newblock \bibinfo{journal}{\emph{arXiv preprint arXiv:2406.04333}} (\bibinfo{year}{2024}).
\newblock


\bibitem[Wang et~al\mbox{.}(2024b)]%
        {wang2024quest}
\bibfield{author}{\bibinfo{person}{Haoxuan Wang}, \bibinfo{person}{Yuzhang Shang}, \bibinfo{person}{Zhihang Yuan}, \bibinfo{person}{Junyi Wu}, {and} \bibinfo{person}{Yan Yan}.} \bibinfo{year}{2024}\natexlab{b}.
\newblock \showarticletitle{Quest: Low-bit diffusion model quantization via efficient selective finetuning}.
\newblock \bibinfo{journal}{\emph{arXiv preprint arXiv:2402.03666}} (\bibinfo{year}{2024}).
\newblock


\bibitem[Wang et~al\mbox{.}(2024a)]%
        {wang2024lavie}
\bibfield{author}{\bibinfo{person}{Yaohui Wang}, \bibinfo{person}{Xinyuan Chen}, \bibinfo{person}{Xin Ma}, \bibinfo{person}{Shangchen Zhou}, \bibinfo{person}{Ziqi Huang}, \bibinfo{person}{Yi Wang}, \bibinfo{person}{Ceyuan Yang}, \bibinfo{person}{Yinan He}, \bibinfo{person}{Jiashuo Yu}, \bibinfo{person}{Peiqing Yang}, {et~al\mbox{.}}} \bibinfo{year}{2024}\natexlab{a}.
\newblock \showarticletitle{Lavie: High-quality video generation with cascaded latent diffusion models}.
\newblock \bibinfo{journal}{\emph{International Journal of Computer Vision}} (\bibinfo{year}{2024}), \bibinfo{pages}{1--20}.
\newblock


\bibitem[Wang et~al\mbox{.}(2024c)]%
        {wang2024lavin}
\bibfield{author}{\bibinfo{person}{Zhaoqing Wang}, \bibinfo{person}{Xiaobo Xia}, \bibinfo{person}{Runnan Chen}, \bibinfo{person}{Dongdong Yu}, \bibinfo{person}{Changhu Wang}, \bibinfo{person}{Mingming Gong}, {and} \bibinfo{person}{Tongliang Liu}.} \bibinfo{year}{2024}\natexlab{c}.
\newblock \showarticletitle{LaVin-DiT: Large Vision Diffusion Transformer}.
\newblock \bibinfo{journal}{\emph{arXiv preprint arXiv:2411.11505}} (\bibinfo{year}{2024}).
\newblock


\bibitem[Watson et~al\mbox{.}(2022)]%
        {watson2022learning}
\bibfield{author}{\bibinfo{person}{Daniel Watson}, \bibinfo{person}{William Chan}, \bibinfo{person}{Jonathan Ho}, {and} \bibinfo{person}{Mohammad Norouzi}.} \bibinfo{year}{2022}\natexlab{}.
\newblock \showarticletitle{Learning fast samplers for diffusion models by differentiating through sample quality}.
\newblock \bibinfo{journal}{\emph{arXiv preprint arXiv:2202.05830}} (\bibinfo{year}{2022}).
\newblock


\bibitem[Wei et~al\mbox{.}(2023)]%
        {wei2023outlier}
\bibfield{author}{\bibinfo{person}{Xiuying Wei}, \bibinfo{person}{Yunchen Zhang}, \bibinfo{person}{Yuhang Li}, \bibinfo{person}{Xiangguo Zhang}, \bibinfo{person}{Ruihao Gong}, \bibinfo{person}{Jinyang Guo}, {and} \bibinfo{person}{Xianglong Liu}.} \bibinfo{year}{2023}\natexlab{}.
\newblock \showarticletitle{Outlier suppression+: Accurate quantization of large language models by equivalent and optimal shifting and scaling}.
\newblock \bibinfo{journal}{\emph{arXiv preprint arXiv:2304.09145}} (\bibinfo{year}{2023}).
\newblock


\bibitem[Wimbauer et~al\mbox{.}(2024)]%
        {wimbauer2024cache}
\bibfield{author}{\bibinfo{person}{Felix Wimbauer}, \bibinfo{person}{Bichen Wu}, \bibinfo{person}{Edgar Schoenfeld}, \bibinfo{person}{Xiaoliang Dai}, \bibinfo{person}{Ji Hou}, \bibinfo{person}{Zijian He}, \bibinfo{person}{Artsiom Sanakoyeu}, \bibinfo{person}{Peizhao Zhang}, \bibinfo{person}{Sam Tsai}, \bibinfo{person}{Jonas Kohler}, {et~al\mbox{.}}} \bibinfo{year}{2024}\natexlab{}.
\newblock \showarticletitle{Cache me if you can: Accelerating diffusion models through block caching}. In \bibinfo{booktitle}{\emph{Proceedings of the IEEE/CVF Conference on Computer Vision and Pattern Recognition}}. \bibinfo{pages}{6211--6220}.
\newblock


\bibitem[Wong et~al\mbox{.}(2024)]%
        {wong2024analysis}
\bibfield{author}{\bibinfo{person}{Kit Wong}, \bibinfo{person}{Rolf Dornberger}, {and} \bibinfo{person}{Thomas Hanne}.} \bibinfo{year}{2024}\natexlab{}.
\newblock \showarticletitle{An analysis of weight initialization methods in connection with different activation functions for feedforward neural networks}.
\newblock \bibinfo{journal}{\emph{Evolutionary Intelligence}} \bibinfo{volume}{17}, \bibinfo{number}{3} (\bibinfo{year}{2024}), \bibinfo{pages}{2081--2089}.
\newblock


\bibitem[Wu et~al\mbox{.}(2024)]%
        {wu2024ptq4dit}
\bibfield{author}{\bibinfo{person}{Junyi Wu}, \bibinfo{person}{Haoxuan Wang}, \bibinfo{person}{Yuzhang Shang}, \bibinfo{person}{Mubarak Shah}, {and} \bibinfo{person}{Yan Yan}.} \bibinfo{year}{2024}\natexlab{}.
\newblock \showarticletitle{PTQ4DiT: Post-training Quantization for Diffusion Transformers}.
\newblock \bibinfo{journal}{\emph{arXiv preprint arXiv:2405.16005}} (\bibinfo{year}{2024}).
\newblock


\bibitem[Xiao et~al\mbox{.}(2023c)]%
        {xiao2023smoothquant}
\bibfield{author}{\bibinfo{person}{Guangxuan Xiao}, \bibinfo{person}{Ji Lin}, \bibinfo{person}{Mickael Seznec}, \bibinfo{person}{Hao Wu}, \bibinfo{person}{Julien Demouth}, {and} \bibinfo{person}{Song Han}.} \bibinfo{year}{2023}\natexlab{c}.
\newblock \showarticletitle{Smoothquant: Accurate and efficient post-training quantization for large language models}. In \bibinfo{booktitle}{\emph{International Conference on Machine Learning}}. PMLR, \bibinfo{pages}{38087--38099}.
\newblock


\bibitem[Xiao et~al\mbox{.}(2023a)]%
        {xiao2023dcifpn}
\bibfield{author}{\bibinfo{person}{Junrui Xiao}, \bibinfo{person}{He Jiang}, \bibinfo{person}{Zhikai Li}, {and} \bibinfo{person}{Qingyi Gu}.} \bibinfo{year}{2023}\natexlab{a}.
\newblock \showarticletitle{DCIFPN: Deformable cross-scale interaction feature pyramid network for object detection}.
\newblock \bibinfo{journal}{\emph{IET Image Processing}} (\bibinfo{year}{2023}).
\newblock


\bibitem[Xiao et~al\mbox{.}(2023b)]%
        {xiao2023patch}
\bibfield{author}{\bibinfo{person}{Junrui Xiao}, \bibinfo{person}{Zhikai Li}, \bibinfo{person}{Lianwei Yang}, {and} \bibinfo{person}{Qingyi Gu}.} \bibinfo{year}{2023}\natexlab{b}.
\newblock \showarticletitle{Patch-wise Mixed-Precision Quantization of Vision Transformer}.
\newblock \bibinfo{journal}{\emph{arXiv preprint arXiv:2305.06559}} (\bibinfo{year}{2023}).
\newblock


\bibitem[Xu et~al\mbox{.}(2024)]%
        {xu2024imagereward}
\bibfield{author}{\bibinfo{person}{Jiazheng Xu}, \bibinfo{person}{Xiao Liu}, \bibinfo{person}{Yuchen Wu}, \bibinfo{person}{Yuxuan Tong}, \bibinfo{person}{Qinkai Li}, \bibinfo{person}{Ming Ding}, \bibinfo{person}{Jie Tang}, {and} \bibinfo{person}{Yuxiao Dong}.} \bibinfo{year}{2024}\natexlab{}.
\newblock \showarticletitle{Imagereward: Learning and evaluating human preferences for text-to-image generation}.
\newblock \bibinfo{journal}{\emph{Advances in Neural Information Processing Systems}}  \bibinfo{volume}{36} (\bibinfo{year}{2024}).
\newblock


\bibitem[Yu et~al\mbox{.}(2015)]%
        {yu2015lsun}
\bibfield{author}{\bibinfo{person}{Fisher Yu}, \bibinfo{person}{Ari Seff}, \bibinfo{person}{Yinda Zhang}, \bibinfo{person}{Shuran Song}, \bibinfo{person}{Thomas Funkhouser}, {and} \bibinfo{person}{Jianxiong Xiao}.} \bibinfo{year}{2015}\natexlab{}.
\newblock \showarticletitle{Lsun: Construction of a large-scale image dataset using deep learning with humans in the loop}.
\newblock \bibinfo{journal}{\emph{arXiv preprint arXiv:1506.03365}} (\bibinfo{year}{2015}).
\newblock


\bibitem[Zhang et~al\mbox{.}(2023a)]%
        {zhang2023dual}
\bibfield{author}{\bibinfo{person}{Luoming Zhang}, \bibinfo{person}{Wen Fei}, \bibinfo{person}{Weijia Wu}, \bibinfo{person}{Yefei He}, \bibinfo{person}{Zhenyu Lou}, {and} \bibinfo{person}{Hong Zhou}.} \bibinfo{year}{2023}\natexlab{a}.
\newblock \showarticletitle{Dual Grained Quantization: Efficient Fine-Grained Quantization for LLM}.
\newblock \bibinfo{journal}{\emph{arXiv preprint arXiv:2310.04836}} (\bibinfo{year}{2023}).
\newblock


\bibitem[Zhang et~al\mbox{.}(2023c)]%
        {zhang2023adding}
\bibfield{author}{\bibinfo{person}{Lvmin Zhang}, \bibinfo{person}{Anyi Rao}, {and} \bibinfo{person}{Maneesh Agrawala}.} \bibinfo{year}{2023}\natexlab{c}.
\newblock \showarticletitle{Adding conditional control to text-to-image diffusion models}. In \bibinfo{booktitle}{\emph{Proceedings of the IEEE/CVF International Conference on Computer Vision}}. \bibinfo{pages}{3836--3847}.
\newblock


\bibitem[Zhang et~al\mbox{.}(2022)]%
        {zhang2022gddim}
\bibfield{author}{\bibinfo{person}{Qinsheng Zhang}, \bibinfo{person}{Molei Tao}, {and} \bibinfo{person}{Yongxin Chen}.} \bibinfo{year}{2022}\natexlab{}.
\newblock \showarticletitle{gDDIM: Generalized denoising diffusion implicit models}.
\newblock \bibinfo{journal}{\emph{arXiv preprint arXiv:2206.05564}} (\bibinfo{year}{2022}).
\newblock


\bibitem[Zhang et~al\mbox{.}(2023b)]%
        {zhang2023inversion}
\bibfield{author}{\bibinfo{person}{Yuxin Zhang}, \bibinfo{person}{Nisha Huang}, \bibinfo{person}{Fan Tang}, \bibinfo{person}{Haibin Huang}, \bibinfo{person}{Chongyang Ma}, \bibinfo{person}{Weiming Dong}, {and} \bibinfo{person}{Changsheng Xu}.} \bibinfo{year}{2023}\natexlab{b}.
\newblock \showarticletitle{Inversion-based style transfer with diffusion models}. In \bibinfo{booktitle}{\emph{Proceedings of the IEEE/CVF Conference on Computer Vision and Pattern Recognition}}. \bibinfo{pages}{10146--10156}.
\newblock


\end{thebibliography}

\newpage
\appendix
\onecolumn
\begin{center}  
{\Large \textbf{DilateQuant: Accurate and Efficient Quantization-Aware Training for Diffusion Models \\ via Weight Dilation}} \\
\vspace{0.3cm}
{\large \textbf{Supplementary Materials}}

\end{center}

\section{Proof of Quantization Error}\label{p:1}
\begin{equation}
    \begin{aligned}
        E(\bm{X}, \bm{W})= & \|\bm{X} \bm{W}-Q(\bm{X}) Q(\bm{W})\|_1 \\
        = & \|\bm{X} \bm{W}-\bm{X} Q(\bm{W})+\bm{X} Q(\bm{W})-Q(\bm{X}) Q(\bm{W})\|_1 \\
        \leq & \|\bm{X}(\bm{W}-Q(\bm{W}))\|_1+\|(\bm{X}-Q(\bm{X})) Q(\bm{W})\|_1 \\
        \leq & \|\bm{X}\|_1\|\bm{W}-Q(\bm{W})\|_1+\|\bm{X}-Q(\bm{X})\|_1\|Q(\bm{W})\|_1 \\
        \leq & \|\bm{X}\|_1\|\bm{W}-Q(\bm{W})\|_1+\|\bm{X}-Q(\bm{X})\|_1\|\bm{W}-(\bm{W}-Q(\bm{W}))\|_1 \\
        \leq & \|\bm{X}\|_1\|\bm{W}-Q(\bm{W})\|_1+\|\bm{X}-Q(\bm{X})\|_1\left(\|\bm{W}\|_1+\|\bm{W}-Q(\bm{W})\|_1\right)
    \end{aligned}
\end{equation}

\section{Detailed Experimental Implementations}\label{app:detail}
In this section, we present detailed experimental implementations, including the pre-training models, qunatization settings, and evaluation.

The pre-training models of DDPM\footnote{\url{https://github.com/ermongroup/ddim}}, LDM\footnote{\url{https://github.com/CompVis/latent-diffusion}}, and DiT-XL/2\footnote{\url{https://github.com/facebookresearch/DiT}} are obtained from the official websites.
For text-conditional generation with Stable-Diffusion, we use the CompVis codebase\footnote{\url{https://github.com/CompVis/stable-diffusion}} and its v1.4 checkpoint.

The LDMs consist of a diffusion model and a decoder model.
Following the previous works~\citep{liu2024enhanced,he2023efficientdm,wang2024quest}, DilateQuant focus only on the diffusion models and does not quantize the decoder models.
We empoly channel-wise asymmetric quantization for weights and layer-wise asymmetric quantization for activations.
The input and output layers of models use a fixed 8-bit quantization, as it is a common practice. The weight and activation quantization ranges are initially determined by minimizing values error, and then optimized by our knowledge distillation strategy to align quantized models with full-precision models at block level. 
Since the two compared methods employ non-standard settings, we modify them to standard settings for a fair comparison.
More specifically, we quantize all layers for EfficientDM, including {\small{\verb|Upsample|}}, {\small{\verb|Skip_connection|}}, and {\small{\verb|AttentionBlock's qkvw|}}, which lack quantization in open-source code\footnote{\url{https://github.com/ThisisBillhe/EfficientDM}}.
However, when these layers, which are important for quantization, are added, the performance of EfficientDM degrades drastically.
To recover performance, we double the number of training iterations.
QuEST utilizes channel-wise quantization for activations at 4-bit precision in the code\footnote{\url{https://github.com/hatchetProject/QuEST}}, which is not supported by hardware. Therefore, we adjust the quantization setting to layer-wise quantization for activations.

For experimental evaluation, we use open-source tool \emph{pytorch-OpCounter}\footnote{\url{https://github.com/Lyken17/pytorch-OpCounter}} to calculate the Model Size and bits operations (Bops) before and after quantization.
And following the quantization settings, we only calculate the diffusion model part, not the decoder and encoder parts.
We use the ADM’s TensorFlow evaluation suite \emph{guided-diffusion}\footnote{\url{https://github.com/openai/guided-diffusion}} to evaluate FID, sFID, and IS, \emph{clip-score}\footnote{\url{https://github.com/Taited/clip-score}} to evaluate CLIP scores, and \emph{Aesthetic Predictor}\footnote{\url{https://github.com/shunk031/simple-aesthetics-predictor}} to evaluate Aesthetic Score. 
As the per practice~\citep{liu2024enhanced,wang2024quest}, we employ the zero-shot approach to evaluate Stable-Diffusion on COCO-val, resizing the generated 512 $\times $ 512 images and validation images in 300 $\times $ 300 with the center cropping to evaluate FID score.
All experiments are performed on one RTX A6000.

\section{Robustness of DilateQuant for Samplers and Time Steps}\label{app:sampler}
To assess the robustness of DilateQuant for samplers, we conduct experiments over LDM-4 on ImageNet with three distant samplers, including DDIMsampler~\cite{song2020denoising}, PLMSsampler~\cite{liu2022pseudo}, and DPMSolversampler~\cite{lu2022dpm}.
Given that time step is the most important hyperparameter for diffusion models, we also evaluate DilateQuant for models with different time steps, including 20 steps and 100 steps.
As shown in Table~\ref{tab:robust}, our method showcases excellent robustness across different samplers and time steps, leading to significant performance enhancements compared to previous methods.
Specifically, our method outperforms the full-precision models in terms of FID and sFID at 6-bit quantization, and the advantages of our method are more pronounced compared to existing methods at the lower 4-bit quantization.

\begin{table}[!t]\small
    \centering
    \caption{The robustness of DilateQuant for time steps and samplers.}
    \vspace{-0.2cm}
    \label{tab:robust}
    \setlength{\tabcolsep}{1.8mm}
    \begin{tabular}{ccccccc}
    \toprule
    Task & Method & Calib. & Prec. (W/A) & FID~$\downarrow$ & sFID~$\downarrow$ & IS~$\uparrow$ \\
    \cmidrule(r){1-7}
    \multirow{7}{3.5cm}{\centering \\ LDM-4 — DDIM \\ time steps = 20} & FP & - & 32/32 & 11.69 & 7.67 & 364.72  \\ 
    \cmidrule(r){2-7}
    \multirow{7}{*}{} & EDA-DM~\textsuperscript{$\dagger$} & 1024 & 6/6 & 11.52 & 8.02 & \textbf{360.77} \\
    \multirow{7}{*}{} & EfficientDM~\textsuperscript{$\star$} & 102.4K & 6/6 & 8.69 & 8.10 & 309.52 \\
    \multirow{7}{*}{} & DilateQuant~ & 1024 & 6/6 & \textbf{8.25} & \textbf{7.66} & 312.30 \\
    \cmidrule(r){2-7}
    \multirow{7}{*}{} & EDA-DM~\textsuperscript{$\dagger$} & 1024 & 4/4 & 20.02 & 36.66 & 204.93 \\
    \multirow{7}{*}{} & EfficientDM~\textsuperscript{$\star$} & 102.4K & 4/4 & 12.08 & 14.75 & 122.12 \\
    \multirow{7}{*}{} & DilateQuant~ & 1024 & 4/4 & \textbf{8.01} & \textbf{13.92} & \textbf{257.24} \\
    \cmidrule(r){1-7}
    \multirow{7}{3.5cm}{\centering \\ LDM-4 — PLMS \\ time steps = 20} & FP & - & 32/32 & 11.71 & 7.08 & 379.19 \\ 
    \cmidrule(r){2-7}
    \multirow{7}{*}{} & EDA-DM~\textsuperscript{$\dagger$} & 1024 & 6/6 & 11.27 & 6.59 & \textbf{363.00} \\
    \multirow{7}{*}{} & EfficientDM~\textsuperscript{$\star$} & 102.4K & 6/6 & 9.85 & 9.36 & 325.13 \\
    \multirow{7}{*}{} & DilateQuant~ & 1024 & 6/6 & \textbf{7.68} & \textbf{5.69} & 315.85 \\
    \cmidrule(r){2-7}
    \multirow{7}{*}{} & EDA-DM~\textsuperscript{$\dagger$} & 1024 & 4/4 & 17.56 & 32.63 & 203.15 \\
    \multirow{7}{*}{} & EfficientDM~\textsuperscript{$\star$} & 102.4K & 4/4 & 14.78 & 9.89 & 103.34 \\
    \multirow{7}{*}{} & DilateQuant~ & 1024 & 4/4 & \textbf{9.56} & \textbf{8.12} & \textbf{243.72} \\
    \cmidrule(r){1-7}
    \multirow{7}{3.5cm}{\centering \\ LDM-4 — DPM-Solver \\ time steps = 20} & FP & - & 32/32 & 11.44 & 6.85 & 373.12 \\ 
    \cmidrule(r){2-7}
    \multirow{7}{*}{} & EDA-DM~\textsuperscript{$\dagger$} & 1024 & 6/6 & 11.14 & 7.95 & \textbf{357.16} \\
    \multirow{7}{*}{} & EfficientDM~\textsuperscript{$\star$} & 102.4K & 6/6 & 8.54 & 9.30 & 336.11 \\
    \multirow{7}{*}{} & DilateQuant~ & 1024 & 6/6 & \textbf{7.32} & \textbf{6.68} & 330.32 \\
    \cmidrule(r){2-7}
    \multirow{7}{*}{} & EDA-DM~\textsuperscript{$\dagger$} & 1024 & 4/4 & 30.86 & 39.40 & 138.01 \\
    \multirow{7}{*}{} & EfficientDM~\textsuperscript{$\star$} & 102.4K & 4/4 & 14.36 & 13.82 & 109.52 \\
    \multirow{7}{*}{} & DilateQuant~ & 1024 & 4/4 & \textbf{8.98} & \textbf{9.97} & \textbf{247.62} \\
    \cmidrule(r){1-7}
    \multirow{7}{3.5cm}{\centering \\ LDM-4 — DDIM \\ time steps = 100} & FP & - & 32/32 & 4.45 & 6.27 & 238.39 \\ 
    \cmidrule(r){2-7}
    \multirow{7}{*}{} & EDA-DM~\textsuperscript{$\dagger$} & 1024 & 6/6 & 12.21 & 12.13 & 71.50 \\
    \multirow{7}{*}{} & EfficientDM~\textsuperscript{$\star$} & 102.4K & 6/6 & \textbf{5.57} & 7.50 & \textbf{165.15} \\
    \multirow{7}{*}{} & DilateQuant~ & 1024 & 6/6 & 5.97 & \textbf{7.44} & 162.93 \\
    \cmidrule(r){2-7}
    \multirow{7}{*}{} & EDA-DM~\textsuperscript{$\dagger$} & 1024 & 4/4 & N/A & N/A & N/A \\
    \multirow{7}{*}{} & EfficientDM~\textsuperscript{$\star$} & 102.4K & 4/4 & 20.70 & 11.79 & 72.67 \\
    \multirow{7}{*}{} & DilateQuant~ & 1024 & 4/4 & \textbf{9.85} & \textbf{10.79} & \textbf{147.63} \\
    \bottomrule
\end{tabular}
    \vspace{-0.3cm}
\end{table}
\begin{table}[!t]\small
    \centering
    \caption{Efficiency comparisons of various quantization frameworks at 4-bit precision.}
    \vspace{-0.2cm}
    \label{tab:efficient}
    \setlength{\tabcolsep}{1.8mm}
    \begin{tabular}{cccccc}
        \toprule 
        Model & Method & Calib. & Time Cost (hours) & GPU Memory (MB) & FID~$\downarrow$ \\ 
        \cmidrule(r){1-6}
        \multirow{3}{2.5cm}{\centering DDPM \\ CIFAR-10} & PTQ & 5120 & 0.97 & 3019 & 120.24 \\ 
        \multirow{3}{*}{} & V-QAT & 819.2K & 2.98 & 9546 & 81.27 \\ 
        \multirow{3}{*}{} & Ours & 5120 & 1.08 & 3439 & 9.13 \\ 
        \cmidrule(r){1-6}
        \multirow{3}{2.5cm}{\centering LDM \\ ImageNet} & PTQ & 1024 & 6.43 & 13831 & 20.02 \\ 
        \multirow{3}{*}{} & V-QAT & 102.4K  & 5.20 & 22746 & 12.08 \\ 
        \multirow{3}{*}{} & Ours & 1024 & 6.56 & 14680 & 8.01 \\ 
        \cmidrule(r){1-6}
        \multirow{3}{2.5cm}{\centering Stable-Diffusion \\MS-COCO} & PTQ & 512 & 7.23 & 30265 & 236.31 \\ 
        \multirow{3}{*}{} & V-QAT & 12.8K & 30.25 & 46082 & 216.43 \\ 
        \multirow{3}{*}{} & Ours & 512 & 7.41 & 31942 & 42.97 \\ 
        \bottomrule
    \end{tabular}
    \vspace{-0.3cm}
\end{table}

\section{Efficiency Comparisons of Various Quantization Frameworks}\label{app:efficient}
We investigate the efficiency of DilateQuant across data resource, time cost, and GPU memory. 
We compare our method with PTQ-based method~\citep{liu2024enhanced} and variant QAT-based method~\citep{he2023efficientdm} on the mainstream diffusion models (DDPM, LDM, Stable-Diffusion). 
As reported in Table~\ref{tab:efficient}, our method performs PTQ-like efficiency, while significantly improving the performance of the quantized models.
This provides an affordable and efficient quantization process for diffusion models.

\section{Thorough Comparison with EfficientDM and QuEST}\label{app:compare}
EfficientDM~\citep{he2023efficientdm} and QuEST~\citep{wang2024quest} are two variant QAT-based methods, which achieve 4-bit quantization of diffusion models with efficiency.
However, both of them are non-standard.
Specifically, EfficientDM preserves some layers at full-precision, notably the {\small{\verb|Upsample|}}, {\small{\verb|Skip_connection|}}, and the matrix multiplication of {\small{\verb|AttentionBlock's qkvw|}}. 
These layers have been demonstrated to have the most significant impact on the quantization of diffusion models in previous works~\citep{shang2023post,li2023q,liu2024enhanced}.
QuEST employs standard channel-wise quantization for weights and layer-wise quantization for activations at 6-bit precision. However, at 4-bit precision, it uses channel-wise quantization for the activations of all {\small{\verb|Conv|}} and {\small{\verb|Linear|}} layers, which is hardly supported by the hardware because it cannot factor the different scales out of the accumulator summation (please see Appendix~\ref{app:mac} for details), leading to inefficient acceleration.

To thoroughly compare DilateQuant with EfficientDM and QuEST, we conduct experiments on  LSUN-church with standard and non-standard quantization settings, as shown in Table~\ref{tab:compare}.
When neglecting these layers that are important for quantization, DilateQuant extremely reduces the FID to 8.68 with 4-bit quantization.
Compared to the standard setting, the performance improvement is more noticeable.
When setting channel-wise quantization for activations, DilateQuant also reduces a 2.84 FID compared with QuEST.
Conclusively, DilateQuant significantly outperforms EfficientDM and QuEST at different quantization precisions for both standard and non-standard settings, which demonstrates the stability and standards of DilateQuant.
\begin{table}[h]\small
    \centering
    \caption{Comparison with EfficientDM and QuEST in both standard and non-standard settings.}
    \vspace{-0.2cm}
    \label{tab:compare}
    \setlength{\tabcolsep}{1.8mm}
    \begin{tabular}{cccccc}
    \toprule 
    Task & Mode & Method & Prec. & Size (MB) & FID~$\downarrow$ \\
    \cmidrule(r){1-6}
    \multirow{13}{2.2cm}{\centering \\LSUN-Church \citep{yu2015lsun} \\256 $\times $ 256\\$ $\\LDM-8\\steps = 100\\eta = 0.0} & - & FP & 32/32 & 1514.5 & 4.06\\ 
    \cmidrule(r){2-6}
    \multirow{13}{*}{} & \multirow{4}{2.8cm}{\centering \textbf{Non-standard}\\Not quantize for all layers} & EfficientDM & 6/6 & 315.0 & 6.29 \\
    \multirow{13}{*}{} & \multirow{4}{*}{} & DilateQuant & 6/6 & 315.0 & \textbf{4.73} \\
    \cmidrule(r){3-6}
    \multirow{13}{*}{} & \multirow{4}{*}{} & EfficientDM & 4/4 & 222.7 & 14.34 \\
    \multirow{13}{*}{} & \multirow{4}{*}{} & DilateQuant & 4/4 & 222.7 & \textbf{8.68} \\
    \cmidrule(r){2-6}
    \multirow{13}{*}{} & \multirow{4}{2.8cm}{\centering \textbf{Standard}\\Quantize for all layers} & EfficientDM & 6/6 & 284.9 & 6.92 \\
    \multirow{13}{*}{} & \multirow{4}{*}{} & DilateQuant & 6/6 & 284.9 & \textbf{5.33} \\
    \cmidrule(r){3-6}
    \multirow{13}{*}{} & \multirow{4}{*}{} & EfficientDM & 4/4 & 190.3 & 15.08 \\
    \multirow{13}{*}{} & \multirow{4}{*}{} & DilateQuant & 4/4 & 190.3 & \textbf{10.10} \\
    \cmidrule(r){2-6}
    \multirow{13}{*}{} & \multirow{2}{2.8cm}{\centering \textbf{Non-standard}\\Channel-wise for A} & QuEST & 4/4 & 190.3 & 11.76 \\
    \multirow{13}{*}{} & \multirow{2}{*}{} & DilateQuant & 4/4 & 190.3 & \textbf{8.94} \\
    \cmidrule(r){2-6}
    \multirow{13}{*}{} & \multirow{2}{2.8cm}{\centering \textbf{Standard}\\Layer-wise for A} & QuEST & 4/4 & 190.3 & 13.03 \\
    \multirow{13}{*}{} & \multirow{2}{*}{} & DilateQuant & 4/4 & 190.3 & \textbf{10.10} \\
    \bottomrule
\end{tabular}
    \vspace{-0.3cm}
\end{table}

\section{Quantization Error of Activation and Weight}\label{app:error}
According to Eq.~\ref{eq:error1} and~\ref{eq:error2}, the quantization errors for activations and weights can be expressed as:
\begin{align}
    \|\bm{X}-Q(\bm{X})\|_1 = \Delta_x \cdot \mathit{E_{x_{round}}} + \mathit{E_{x_{clip}}}, \quad
    \|\bm{W}-Q(\bm{W})\|_1 = \Delta_w \cdot \mathit{E_{w_{round}}} + \mathit{E_{w_{clip}}} 
\end{align}
In Table~\ref{tab:error}, $\mathit{E_{clip}}$ and $\Delta \cdot \mathit{E_{round}}$ represent the errors caused by the clip and round functions across all layers of the model when generating a single image at 4-bit precision. To eliminate random errors, we set the batch size to 256 for CIFAR-10, LSUN, and ImageNet, set the batch size to 512 for MSCOCO, and then compute the average.
Considering that the errors introduced by the clipping function are minimal, we simplify the quantization errors in this paper as:
\begin{align}
    \|\bm{X}-Q(\bm{X})\|_1 = \Delta_x \cdot \mathit{E_{x_{round}}}, \quad
    \|\bm{W}-Q(\bm{W})\|_1 = \Delta_w \cdot \mathit{E_{w_{round}}} 
\end{align}
\begin{table}[h]\small
    \centering
    \caption{Statistics on quantization errors for different tasks.}
    \vspace{-0.2cm}
    \label{tab:error}
    \setlength{\tabcolsep}{1.8mm}
    \begin{tabular}{cccccc}
        \toprule 
        {\centering Tasks} & CIFAR-10 & LSUN-Bedroom & LSUN-Church & ImageNet & MSCOCO  \\
        \midrule
        $\mathit{E_{x_{clip}}}$ & 4.5\% & 5.3\% & 4.7\% & 5.1\% & 3.5\% \\
        $\Delta_x \cdot \mathit{E_{x_{round}}}$ & 95.5\% & 94.7\% & 95.3\% & 94.9\% & 96.5\% \\ 
        $\|\bm{X}-Q(\bm{X})\|_1 $ & 3.48M & 4.93M & 4.08M & 5.21M & 6.57M \\ 
        \midrule
        $\mathit{E_{w_{clip}}}$ & 2.5\% & 2.1\% & 2.4\% & 2.4\% & 3.0\% \\ 
        $\Delta_w \cdot \mathit{E_{w_{round}}}$ & 97.5\% & 97.9\% & 97.6\% & 97.6\% & 97.0\% \\ 
        $\|\bm{W}-Q(\bm{W})\|_1$ & 4.08K & 7.51K & 5.12K & 8.65K & 10.95K \\ 
        \bottomrule
    \end{tabular}
    \vspace{-0.3cm}
\end{table}

\section{Workflow and Effects of Weight Dilation}\label{app:wd}
The comprehensive workflow of Weight Dilation is illustrated in Algorithm~\ref{alg:algorithm}. We implement WD in three steps: searching unsaturated channels for scaling (Lines 2-3), calculating scaling factor (Lines 5-10), and scaling activations and weights (Line 12).
WD alleviates the wide range activations for diffusion models through a novel equivalent scaling algorithm.
In addition, all operations of WD can be implemented simply, making it efficient.

\begin{algorithm}[th]
    \caption{: Overall workflow of WD}
    \label{alg:algorithm}
    \leftline{\textbf{Input}: full-precision $\bm{X} \in \mathbb{R}^{N\times C_i}$ and $\bm{W} \in \mathbb{R}^{C_i\times C_o}$}
    \leftline{\textbf{Output}: scaled $\bm{X^{'}}$ and $\bm{W^{'}}$}
    \begin{algorithmic}[1] 
        \STATE \textbf{searching unsaturated channels for scaling:}
        \STATE \hspace{1.5em}obtain $\bm{W}_{max} \in \mathbb{R}^{C_o}$ and $\bm{W}_{min} \in \mathbb{R}^{C_o}$
        \STATE \hspace{1.5em}record in-channel indexes of $\bm{W}_{max}$ and $\bm{W}_{min}$ as set $\mathbb{A}$
        \STATE \textbf{calculating scaling factor:}
        \STATE \hspace{1.5em}\textbf{for} $k = 1$ to $C_i$ \textbf{do}
        \STATE \hspace{1.5em}\hspace{1.5em}\textbf{if} $k \in \mathbb{A}$: 
        \STATE \hspace{1.5em}\hspace{1.5em}\hspace{1.5em} set $\bm{s}_k=1$
        \STATE \hspace{1.5em}\hspace{1.5em}\textbf{else}: 
        \STATE \hspace{1.5em}\hspace{1.5em}\hspace{1.5em} calculate scaling factor $\bm{s}_k$ with $\bm{W}_{max}$ and $\bm{W}_{min}$ as constraints
        \STATE \hspace{1.5em}\textbf{end for}
        \STATE \textbf{scaling $\bm{X}$ and $\bm{W}$:}
        \STATE \hspace{1.5em}calculate $\bm{X^{'}} = \bm{X} \;/\; \bm{s}$ and $\bm{W^{'}} = \bm{W} \cdot \bm{s}$
        \STATE \textbf{return} $\bm{X^{'}}$ and $\bm{W^{'}}$
    \end{algorithmic}
\end{algorithm}

We evaluate the effectiveness of WD in a fine-grained manner across different tasks. As reported in Table~\ref{tab:wd}, WD stably maintains $\Delta_x^{'} < \Delta_x$ and $\Delta_w^{'} \approx \Delta_w$, indicating that the activation range is effectively reduced while the weight range remains almost unchanged. The \textit{proportion of $s > 1$} represents the proportion of unsaturated in-channel weights.
\begin{table}[h]\small
    \centering
    \caption{Effects of WD on different tasks with 4-bit quantization.}
    \vspace{-0.2cm}
    \label{tab:wd}
    \setlength{\tabcolsep}{1.8mm}
    \begin{tabular}{llllll}
        \toprule 
        {\centering Tasks} & CIFAR-10 & LSUN-Bedroom & LSUN-Church & ImageNet & MSCOCO  \\
        \midrule
        proportion of $s > 1$ & 39.2\% & 52.4\% & 32.8\% & 36.5\% & 43.8\% \\ 
        $\Delta_x^{'} / \Delta_x$ & 0.91 & 0.92 & 0.91 & 0.92 & 0.90 \\ 
        $\Delta_w^{'} / \Delta_w$ & 1.02 & 1.02 & 1.01 & 1.01 & 1.02 \\ 
        \bottomrule
    \end{tabular}
    \vspace{-0.1cm}
\end{table}

\begin{figure}[!t]
    \centering
    \includegraphics[width=1.0\textwidth]{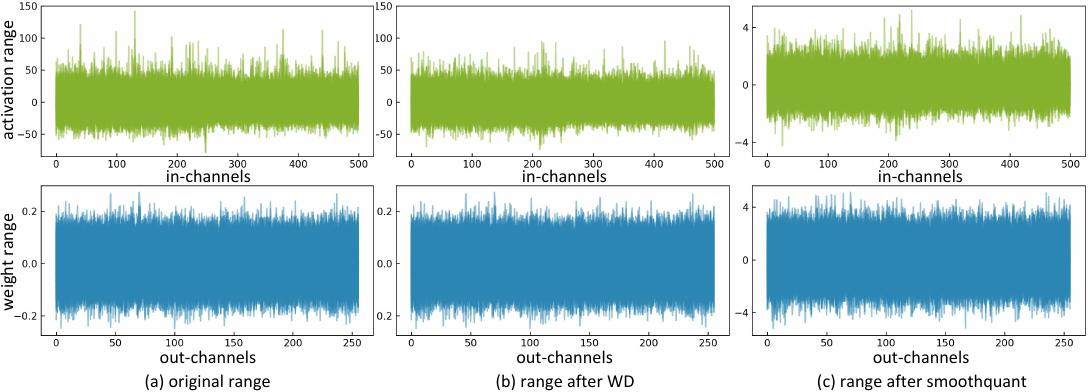}
    \vspace{-0.5cm}
    \caption{Visualization of activation and weight ranges across different scaling methods. The average magnitude of activations across all channels decreases from 51.23 to 45.72 before and after WD.}
    \label{fig:scaling}
    \vspace{-0.2cm}
\end{figure}

\begin{table}[!t]\small 
    \centering
    \caption{The results of various equivalent scaling algorithms for DDIM on CIFAR-10.}
    \vspace{-0.2cm}
    \label{tab:scaling}
    \setlength{\tabcolsep}{1.5mm}
    \begin{tabular}{c | lccccccc}
        \toprule 
        {\centering Prec.} & {\centering Metrics} & No scaling & SmoothQuant & OS+ & OmniQuant & AWQ & DGQ & Ours  \\
        \midrule
        \multirow{4}{0.8cm}{\centering \\ W4A4 } & proportion of $s > 1$ & 0\% & 100\% & 100\% & 100\% & 1\% & 0.5\% & 39.2\% \\ 
        \multirow{4}{*}{} & FID~$\downarrow$ & 9.63 & 9.99 & 9.78 & 9.86 & 10.34 & 9.72 & \textbf{9.13} \\ 
        \multirow{4}{*}{} & sFID~$\downarrow$ & 7.08 & 7.29 & 7.23 & 7.34 & 7.53 & 7.78 & \textbf{6.92} \\ 
        \multirow{4}{*}{} & IS~$\uparrow$ & 8.45 & 8.46 & 8.36 & 8.50 & 8.38 & 8.52 & \textbf{8.56} \\ 
        \midrule
        \multirow{4}{0.8cm}{\centering \\ W6A6 } & proportion of $s > 1$ & 0\% & 100\% & 100\% & 100\% & 1\% & 0.5\% & 39.2\% \\ 
        \multirow{4}{*}{} & FID~$\downarrow$ & 5.75 & 5.44 & 5.81 & 5.56 & 5.85 & 5.09 & \textbf{4.46} \\ 
        \multirow{4}{*}{} & sFID~$\downarrow$ & 4.96 & 4.87 & 4.99 & 4.89 & 5.19 & 4.84 & \textbf{4.64} \\ 
        \multirow{4}{*}{} & IS~$\uparrow$ & 8.80 & 8.86 & 8.76 & 8.81 & 8.78 & 8.89 & \textbf{8.92} \\ 
        \bottomrule
    \end{tabular}  
    \vspace{-0.3cm}
\end{table}

\section{Different Equivalent Scaling Algorithms for Diffusion Models}\label{app:scaling}
\begin{wrapfigure}{r}{5.3cm}
    \centering
    \vspace{-0.2cm}
    \includegraphics[width=0.30\textwidth]{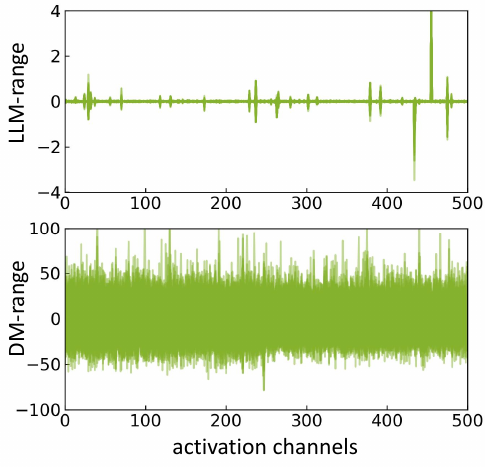}
    \vspace{-0.3cm}
    \caption{The distribution of activation values for LLM (LLaMa-7B) and DM (DDIM).}
    \label{fig:channel}
    \vspace{-0.4cm}
\end{wrapfigure}

In this section, we start by analyzing the differences between LLMs and diffusion models in terms of the challenges of activation quantization.
As shown in Figure~\ref{fig:channel}, the outliers of the diffusion models are present in all channels, unlike in LLMs where the outliers only exist in fixed channels. 
Additionally, the range of activations for diffusion models is also larger than that of the LLMs.
Some equivalent scaling algorithms of PTQ methods are proposed to smooth out the outliers in LLMs, and these methods have achieved success.
SmoothQuant~\citep{xiao2023smoothquant} scales all channels using a hand-designed scaling factor. OS+~\cite{wei2023outlier} conducts channel-wise shifting and scaling across all channels. OmniQuant~\citep{shao2023omniquant} proposes a learnable equivalent transformation to optimize the scaling factors in a differentiable manner.
AWQ~\citep{lin2024awq} only scales a few of channels based on the salient weight. DGQ~\citep{zhang2023dual} devises a percentile scaling scheme to select the scaled channels and calculate the scaling factors. 

Unfortunately, when we applied these previous equivalent scaling algorithms to QAT framework of diffusion models, we find that none of them work.
Specifically, we employ these five methods for diffusion models as follows: 
(1) For SmoothQuant, we scale all channels before quantization using a smoothing factor $\alpha=0.5$; 
(2) For OS+, we perform shifting and scaling across all channels, consistent with the original work;
(3) For OmniQuant, we modify the scaling factors to be learnable variants and train them block by block with a learning rate of 1e-5; 
(4) For AWQ, we scale 1\% of channels based on the salient weight, setting smoothing factor the same as SmoothQuant; 
(5) For DGQ, we scale the top 0.5\% of quantization-sensitive channels, setting scaling factor based on the clipping threshold.
However, as shown in Table~\ref{tab:scaling}, all of these methods result in higher FID and sFID scores compared to no scaling at 4-bit precision.
The reason for this result is that although the range of activations decreases, the range of weights also increases significantly (as shown in Figure~\ref{fig:scaling}), resulting in no change in overall errors. Additionally, the excessive disruption of the original weight range makes models more difficult to converge during the QAT training.
In contrast, the Weight Dilation algorithm searches for unsaturated in-channel weights and dilates them to a constrained range based on the max-min values of the out-channel weights.
The algorithm reduces the range of activations while maintaining the weights range unchanged. This effectively reduces the overall quantization errors and ensures model convergence, reducting the FID and sFID scores to 9.13 and 6.92 at 4-bit precision, respectively.

\section{Hardware-Friendly Quantization}\label{app:mac}
In this section, we investigate the correlation between quantization settings and hardware acceleration.
We start with the principle of quantization to achieve hardware acceleration.
A matrix-vector multiplication, $y = Wx+b$, is calculated by a neural network accelerator, which comprises two fundamental components: the processing elements $C_{n,m}$ and the accumulators $A_n$.
The calculation operation of accelerator is as follows: firstly, the bias values $b_n$ are loaded into accumulators; secondly, the weight values $W_{n,m}$ and the input values $x_m$ are loaded into $C_{n,m}$ and computed in a single cycle; finally, their results are added in the accumulators.
The overall operation is also referred to as Multiply-Accumulate (MAC):
\begin{align}
    A_n = \sum_{m}^{} W_{n,m}x_m + b_n
\end{align}
where $n$ and $m$ represent the out-channel and in-channel of the weights, respectively. 
The pre-trained models are commonly trained using FP32 weights and activations.
In addition to MAC calculations, data needs to be transferred from memory to the processing units.
Both of them severely impact the speed of inference.
Quantization transforms floating-point parameters into fixed-point parameters, which not only reduces the amount of data transfer but also the size and energy consumption of the MAC operation.
This is because the cost of digital arithmetic typically scales linearly to quadratically with the number of bits, and fixed-point addition is more efficient than its floating-point counterpart.
Quantization approximates a floating-point tensor $\bm{x}$ as:
\begin{align}
    \hat{\bm{x}}=\Delta \cdot \bm{x}_{int} \approx \bm{x}
\end{align}

\begin{figure}[!h]
    \centering
    \includegraphics[width=0.65\textwidth]{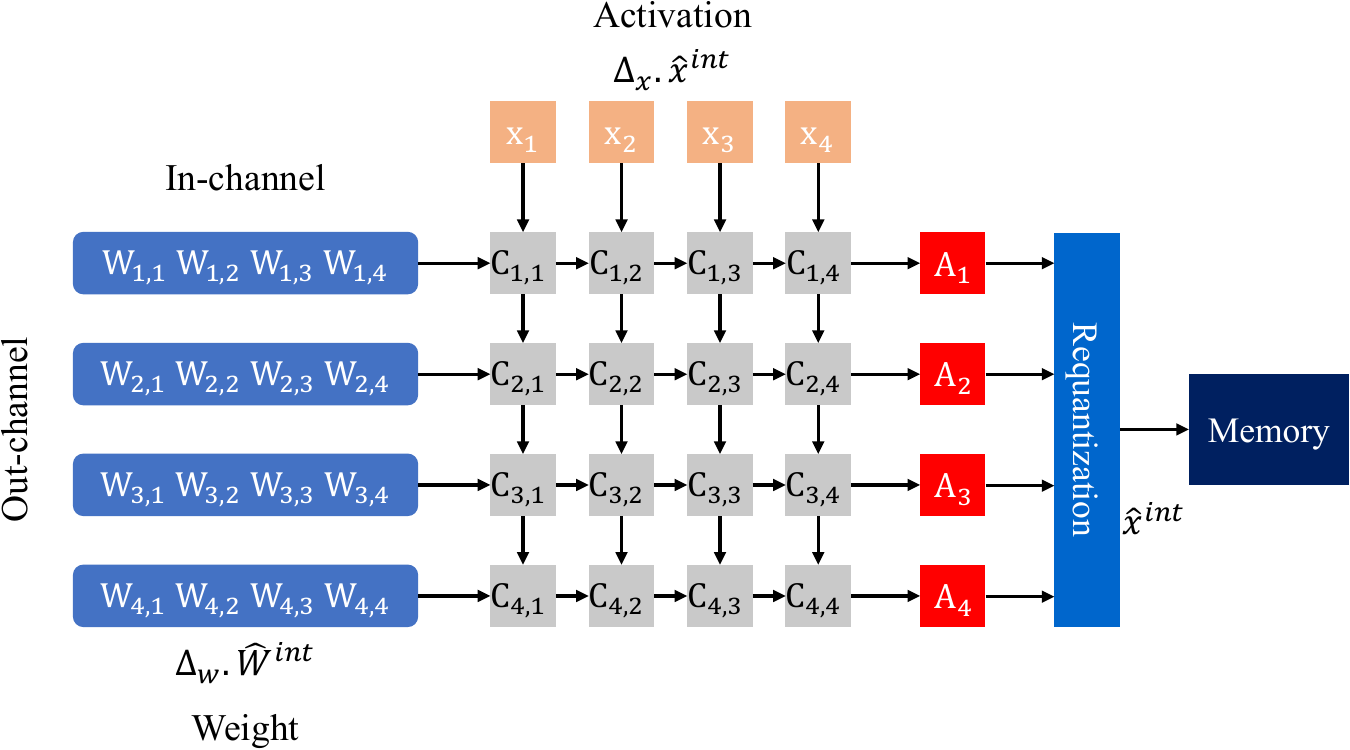}
    \vspace{-0.2cm}
    \caption{A schematic of matrix-multiply logic in accelerator for quantized inference.}
    \label{fig:mac}
    \vspace{-0.3cm}
\end{figure}
where $\bm{x}_{int}$ and $\hat{\bm{x}}$ are integer tensors and quantized tensors, respectively, and $\Delta$ is scale factor.
Quantization settings have different granularity levels. Figure~\ref{fig:mac} shows the accelerator operation after the introduction of quantization. If we set both activations and weights to be layer-wise quantization, the new MAC operation can be represented as:
\begin{align}\label{eq:new MAC}
    \hat{A_n} & = \sum_{m}^{} \hat{W}_{n,m}\hat{x}_m \nonumber + b_n \\
    & = \sum_{m}^{}(\Delta_w \hat{W}_{n,m}^{int})(\Delta_x \hat{x}_m^{int}) + b_n \nonumber \\
    & = \Delta_w \Delta_x \sum_{m}^{}\hat{W}_{n,m}^{int} \hat{x}_m^{int} + b_n 
\end{align}
where $\Delta_w$ and $\Delta_x$ are scale factors for weights and activations, respectively, $\hat{W}_{n,m}^{int}$ and $\hat{x}_m^{int}$ are integer values.
The bias is typically stored in higher bit-width (32-bits), so we ignore bias quantization for now.
As can be seen, this scheme factors out the scale factors from the summation and performs MAC operations in fixed-point format, which accelerates the calculation process.
The activations are quantized back to integer values $\hat{x}_n^{int}$ through a requantization step, which reduces data transfer and simplifies the operations of the next layer.

To approximate the operations of quantization to full-precision, channel-wise quantization for weights is widely used, which sets quantization parameters to each out-channel.
With this setting, the MAC operation in Eq.~\ref{eq:new MAC} can be represented as:
\begin{align}
    \hat{A_n} & = \sum_{m}^{}(\Delta_{w_n} \hat{W}_{n,m}^{int})(\Delta_x \hat{x}_m^{int}) + b_n \nonumber \\
    & = \Delta_{w_n} \Delta_x \sum_{m}^{}\hat{W}_{n,m}^{int} \hat{x}_m^{int} + b_n 
\end{align}
where $\Delta_{w_n}$ is scale factor for the $n_{th}$ out-channel of weights.
However, the channel-wise quantization for activations sets quantization parameters to each in-channel.
This setting is hardly supported by hardware, as the MAC operation is performed as follows: 
\begin{align}
    \hat{A_n} & = \sum_{m}^{}(\Delta_w \hat{W}_{n,m}^{int})(\Delta_{x_m} \hat{x}_m^{int}) + b_n \nonumber \\
    & = \Delta_w \sum_{m}^{}\Delta_{x_m} \hat{W}_{n,m}^{int} \hat{x}_m^{int} + b_n 
\end{align}
where $\Delta_{x_m}$ is scale factor for the $m_{th}$ in-channel of activations.
Due to its inability to factor out the different scales from the accumulator summation, it is not hardware-friendly, leading to invalid acceleration.

\section{Random Samples}\label{app:human}
\begin{figure}[!h]
    \centering
    \includegraphics[width=0.98\textwidth]{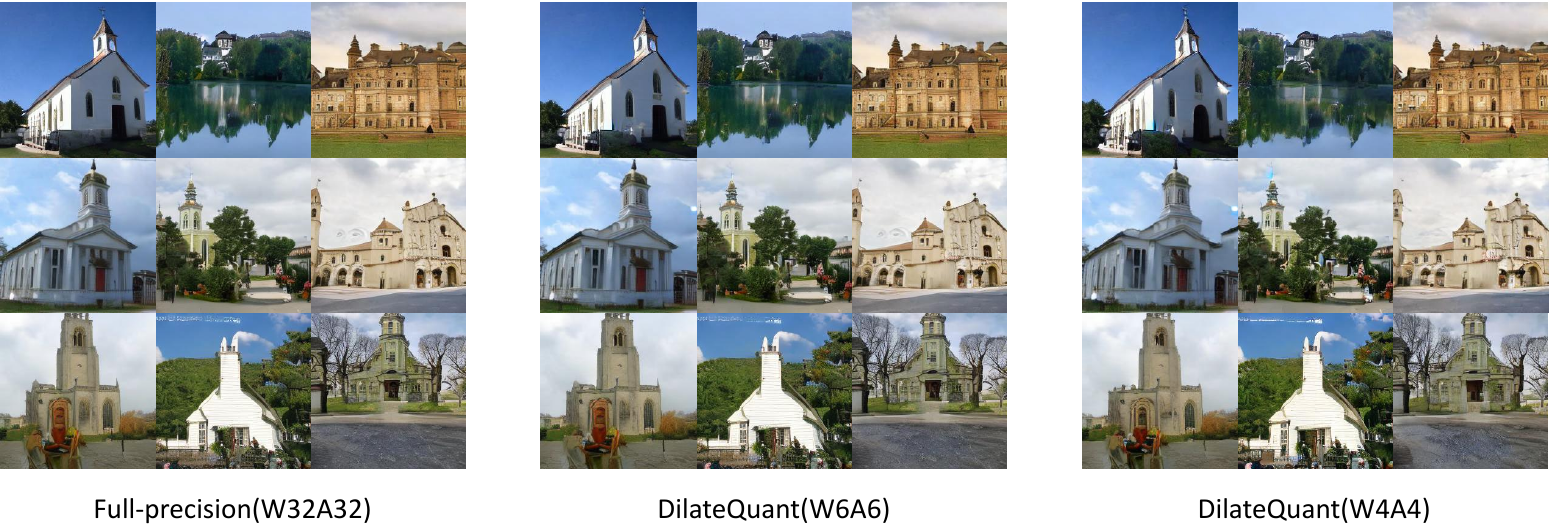}
    \caption{Random samples of quantized models with DilateQuant on LSUN-Church.}
    \label{fig:church}
\end{figure}

\begin{figure}[!h]
    \centering
    \includegraphics[width=0.98\textwidth]{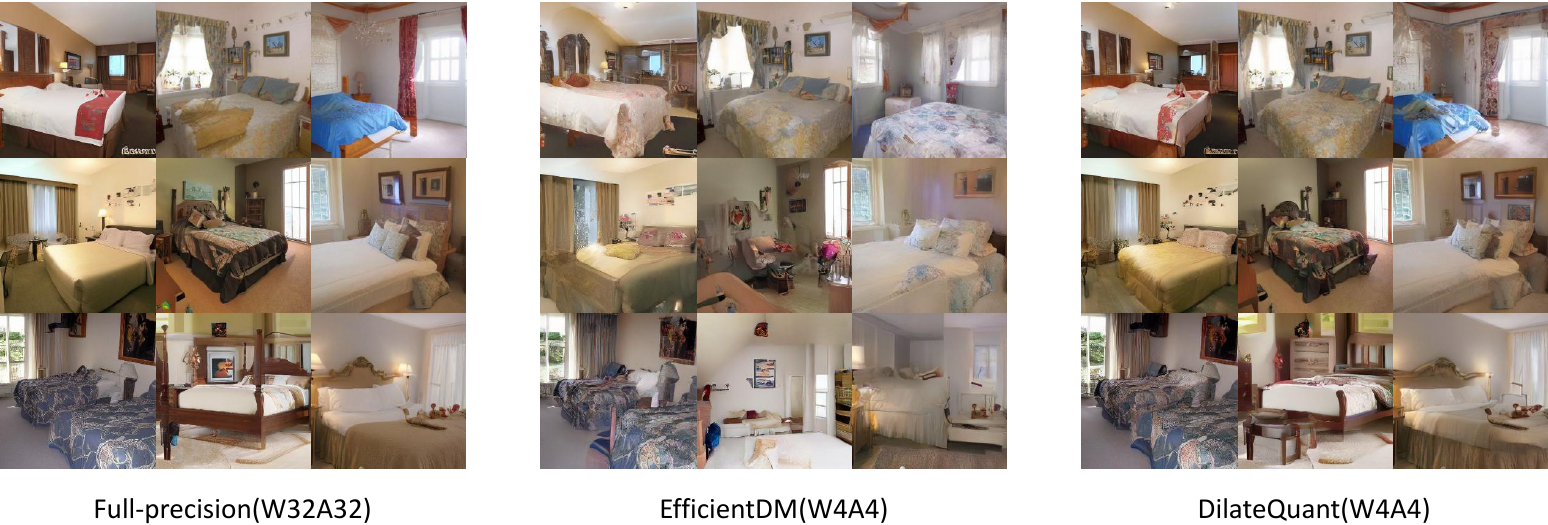}
    \caption{Random samples of different quantized models on LSUN-Bedroom with 4-bit quantization.}
    \label{fig:bedroom}
\end{figure}

\begin{figure}[!h]
    \centering
    \includegraphics[width=0.98\textwidth]{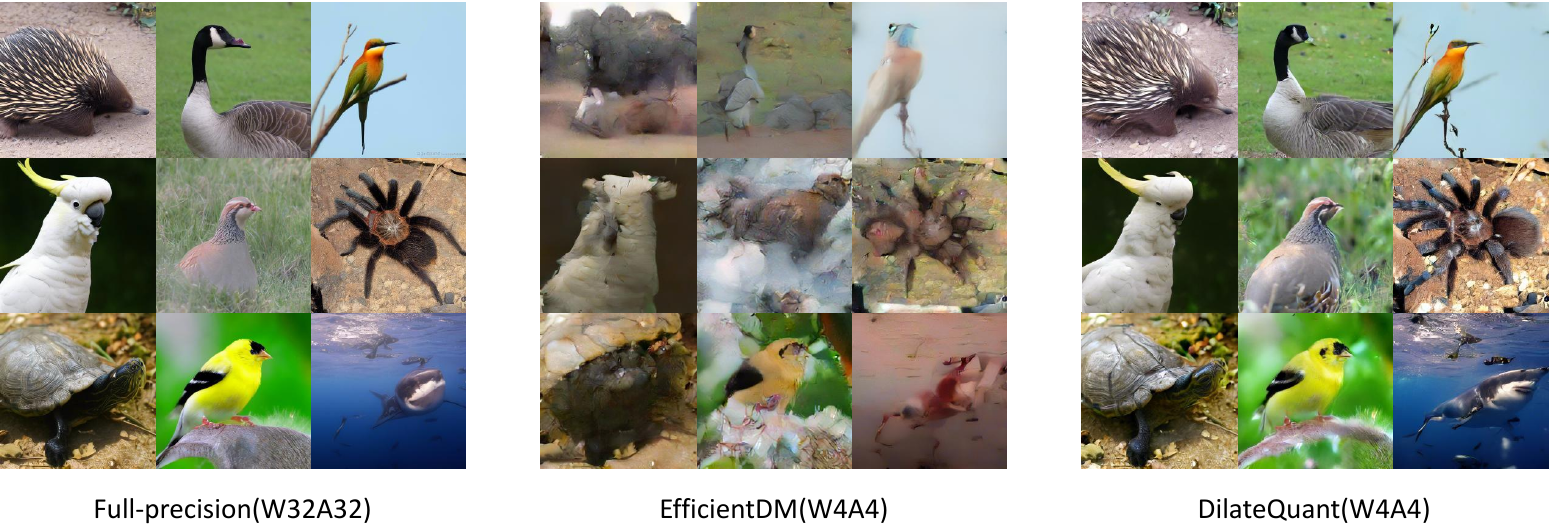}
    \caption{Random samples of different quantized models on ImageNet with 4-bit quantization.}
    \label{fig:imagenet}
\end{figure}

\begin{figure}[!h]
    \centering
    \includegraphics[width=0.8\textwidth]{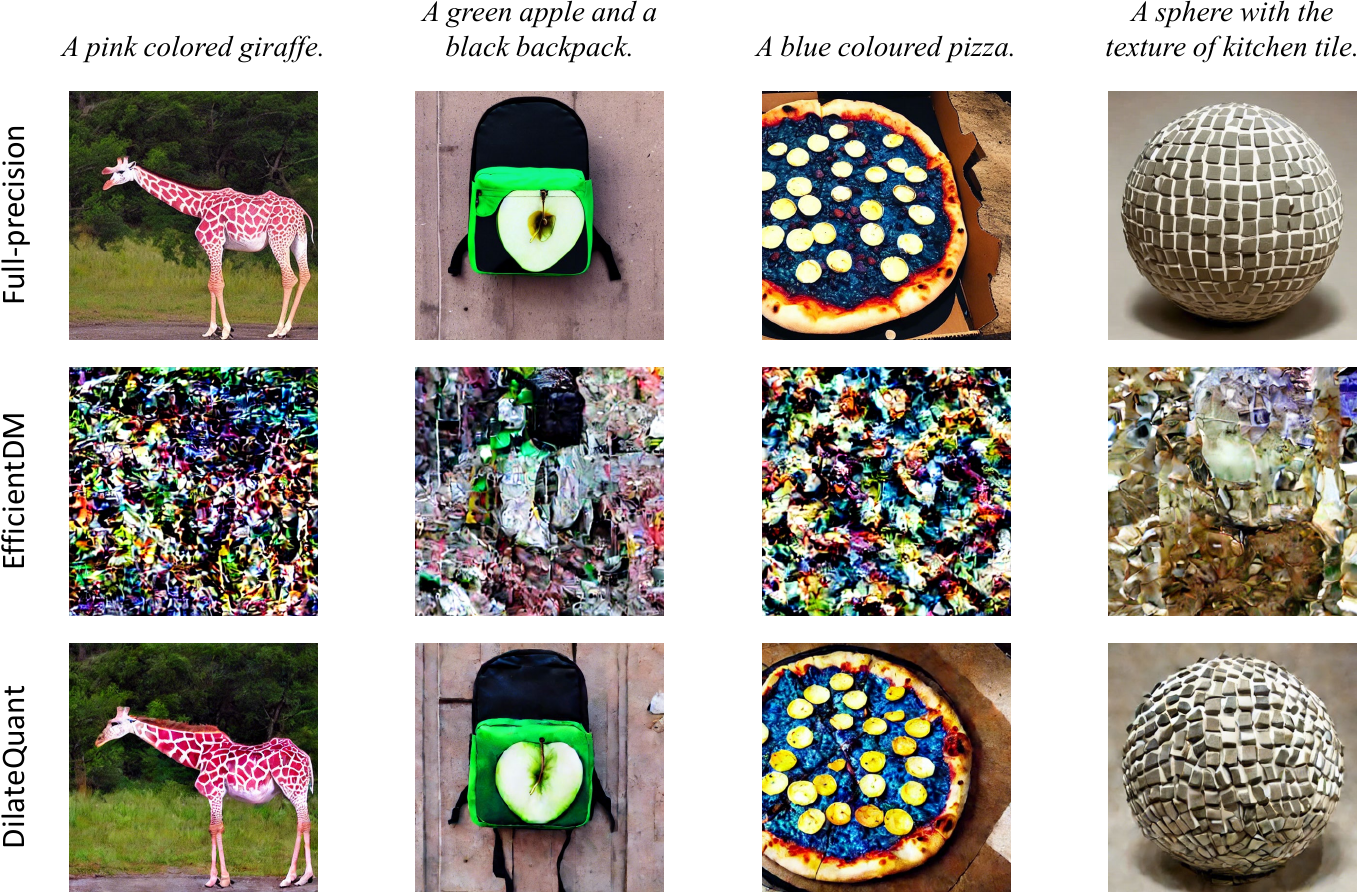}
    \caption{Random samples of different quantized models on DrawBench with 6-bit quantization.}
    \label{fig:drawbench}
\end{figure}

In this section, we visualize the random samples of quantization results in Figure~\ref{fig:church} (LSUN-church),~\ref{fig:bedroom} (LSUN-Bedroom), and~\ref{fig:imagenet} (ImageNet).
For Stable-Diffusion, we use prompts from the convincing DrawBench benchmark to sample, as shown in Figure~\ref{fig:drawbench}.
As can be seen, DilateQuant outperforms previous methods in terms of image quality, fidelity, and diversity.


\end{document}